\documentclass[conference]{IEEEtran}

\usepackage{url}
\usepackage{booktabs}
\usepackage{colortbl}
\usepackage{hyperref}
\usepackage{arydshln}
\usepackage{amsmath}
\usepackage[nameinlink,capitalise]{cleveref}
\usepackage{multirow}
\definecolor{newcolor}{rgb}{.8,.349,.1}
\usepackage{graphicx}
\usepackage{authblk}

\newcommand\blfootnote[1]{%
  \begingroup
  \renewcommand\thefootnote{}\footnote{#1}%
  \addtocounter{footnote}{-1}%
  \endgroup
}

\begin{document}
%
\title{NuClick: A Deep Learning Framework for Interactive Segmentation of Microscopy Images}


\author[1,2]{Navid Alemi Koohbanani*}
\author[3]{Mostafa Jahanifar*}
\author[4]{Neda Zamani Tajadin}
\author[1,2]{, and Nasir Rajpoot}

\affil[1]{Department of Computer Science, University of Warwick, UK}
\affil[2]{Alan Turing Institute, UK}
\affil[3]{Department of Research and Development, NRP co., Iran}
\affil[4]{Department of Electrical Engineering, Tarbiat Modares University, Iran}

\maketitle


\begin{abstract}
Object segmentation is an important step in the workflow of computational pathology. Deep learning based models generally require large amount of labeled data for precise and reliable prediction. However, collecting labeled data is expensive because it often requires expert knowledge, particularly in medical imaging domain where labels are the result of a time-consuming analysis made by one or more human experts. As nuclei, cells and glands are fundamental objects for downstream analysis in computational pathology/cytology, in this paper we propose a simple CNN-based approach to speed up collecting annotations for these objects which requires minimum interaction from the annotator. We show that for nuclei and cells in histology and cytology images, one click inside each object is enough for NuClick to yield a precise annotation. For multicellular structures such as glands, we propose a novel approach to provide the NuClick with a squiggle as a guiding signal, enabling it to segment the glandular boundaries. These supervisory signals are fed to the network as auxiliary inputs along with RGB channels. With detailed experiments, we show that NuClick is adaptable to the object scale, robust against variations in the user input, adaptable to new domains, and delivers reliable annotations. An instance segmentation model trained on masks generated by NuClick achieved the first rank in LYON19 challenge. As exemplar outputs of our framework, we are releasing two datasets: 1) a dataset of lymphocyte annotations within IHC images, and 2) a dataset of segmented WBCs in blood smear images.
\end{abstract}


\section{Introduction}
\blfootnote{* First authors contributed equally.}
Automated analysis of microscopic images heavily relies on classification or segmentation of objects in the image. Starting from a robust and precise segmentation algorithm, downstream analysis subsequently will be more accurate and reliable.
 Deep learning (DL) approaches nowadays have state-of-the-art performance in nearly all computer vision tasks (\cite{russakovsky2015imagenet}). In medical images or more specifically in computational pathology (CP), DL plays an important role for tackling wide range of tasks.
  Despite their success, DL methods have a major problem-their data hungry nature. If they are not provided with sufficient data,  they can easily over-fit on the training data, leading to poor performance on the new unseen data. In computational pathology, most models are trained on datasets that are acquired from just a small sample size of whole data distribution. These models would fail if they are applied on a new distribution (e.g new tissue types or different center that data is coming from). Hence,  one needs to collect annotation from new distribution and then add it to training set to overcome false predictions.
 
 Obtaining annotation as a target for training deep supervised models is time consuming, labour-intensive and sometimes involves expert knowledge. Particularly, for segmentation task where dense annotation is required. It is worth mentioning that in terms of performance, semi-supervised 
 and weakly supervised 
 methods are still far behind fully supervised methods (\cite{taghanaki2019deep}). Therefore, if one needs to build a robust and applicable segmentation algorithm, supervised methods are priority.
 In CP, fully automatic approaches which do not require user interactions have been extensively applied on histology images for segmentation of different objects (e.g. cells, nuclei, glands, etc.) where DL models have shown state-of-the-art performance  (\cite{sirinukunwattana2017gland, kumar2019multi, graham2019hover, koohbanani2019nuclear, pinckaers2019neural, graham2019mild, chen2016dcan, gamper2020pannuke, zhou2019cia}).
Semi-automatic (interactive) segmentation approaches which require the user to provide an input to the system bring several advantages over fully automated approaches: 1) due to the supervisory signal as a prior to the model, interactive models lead to better performance; 2) possible mistakes can be recovered by user interactions; 3) interactive models are less sensitive to domain shift since the supervisory signal can compensate for variations in domains, in other words, interactive models are more generalizable; and 4) selective attribute of interactive models gives the flexibility to the user to choose the arbitrary instances of objects in the visual field (e.g selecting one nucleus for segmentation out of hundreds of nuclei in the ROI).

Due to generalizability power, these models can also serve as annotation tool to facilitate and speed up the annotation collection. Then these annotations can be used to train a fully automatic method for extracting the relevant feature for the task in hand.
For example delineating boundaries of all nuclei, glands or any object of interest is highly labour intensive and time consuming. To be more specific, considering that annotation of one nuclei takes  10s, a visual field containing 100 nuclei takes ~17 minutes to be annotated.
To this end, among interactive models, approaches that require minimum user interaction are of high importance, as it not only minimizes the user effort but also  speed up the process.

In this paper, by concentrating on keeping user interactions as minimum as possible , we propose a unified CNN-based framework for interactive annotation of important microscopic object in three different levels (nuclei, cells, and glands). Our model accepts minimum user interaction which is suitable for collecting annotation in histology domain. 


 \begin{figure}[h]
 \centering
\includegraphics[width =0.9\columnwidth]{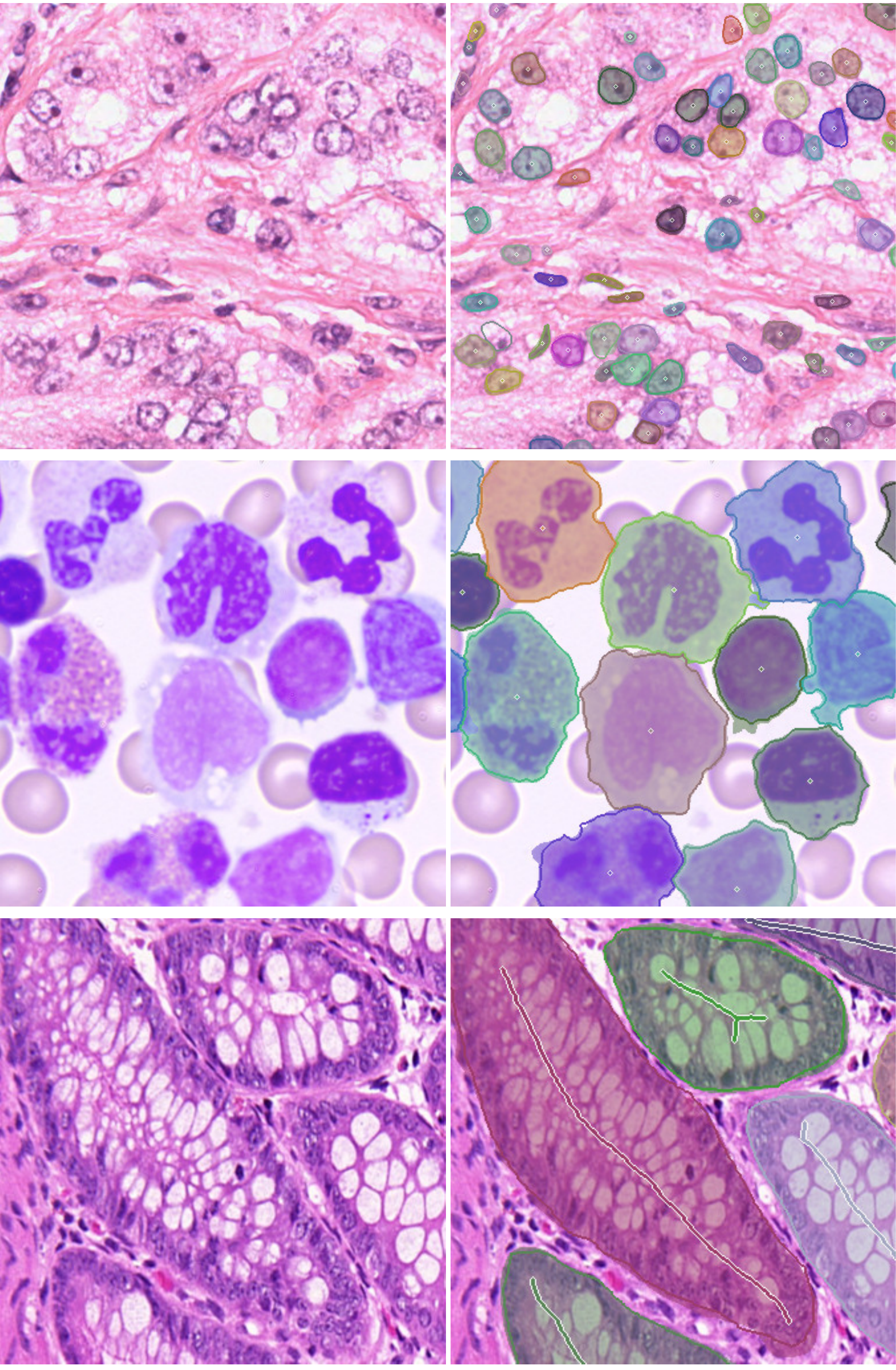}
\caption{\textbf{NuClick interactive segmentation} of objects in histopathological images with different levels of complexity: nuclei (first row), cells (second row), and glands (third row). Solid stroke line around each object outlines the ground truth boundary for that object,  overlaid transparent mask is the predicted segmentation region by NuClick, and points or squiggles indicate the provided guiding signal for interactive segmentation.}
\label{fig:validation}
\end{figure}

 \section{Related Works}
 
\subsection{Weakly Supervised Signals for Segmentation}
Numerous methods have been proposed in the literature that utilise weak labels as supervisory signals. In these methods, supervisory signal serves as an incomplete (weak) ground truth segmentation in the model output. Therefore, a desirable weakly supervised model would be a model that generalize well on the partial supervisory signals and outputs a more complete segmentation of the desired object. These methods are not considered as interactive segmentation methods and are particularly useful when access to full image segmentation labels is limited.
 
For instance, \cite{yoo2019pseudoedgenet} and \cite{qu2019weakly} introduced weakly supervised nucleus segmentation models which are trained based on nuclei centroid points instead of full segmentation masks. Several other works used image-level labels (\cite{pathak2014fully, kolesnikov2016seed, pathak2015constrained, wei2018revisiting}), boxes (\cite{khoreva2017simple}), noisy web labels (\cite{jin2017webly, ahmed2014semantic}), point-clicks (\cite{bearman2016s, bell2015material, chen2018tap, wang2014touchcut}), and  squiggles (\cite{lin2016scribblesup, xu2015learning}) as weak labels to supervise their segmentation models. Our model is analogous to methods proposed by \cite{bearman2016s} and \cite{lin2016scribblesup} with the difference that we used points and squiggles as auxiliary guiding signals in the input of our model. Our model is fully supervised and we will show how this additional information can be used to further improve accuracy of segmentation networks on histology images.

\subsection{Interactive segmentation} \label{Sec-inter}
  Interactive segmentation of objects has been studied for over a decade now. In many works (\cite{bai2009geodesic, batra2011interactively, boykov2001interactive, rother2004grabcut, cheng2015densecut, gulshan2010geodesic, shankar2015video, mortensen1998interactive, cagnoni1999genetic, de2004interactive, wang2018interactive, li2018interactive})  object segmentation is formulated as energy minimization on a graph defined over objects. In a recent unsupervised approach proposed by \cite{papadopoulos2017extreme}, the annotator clicks on  four extreme points (left-most, right-most, top and bottom pixels), then an edge detection algorithm is applied to the whole image to extract boundaries, afterwards the shortest path between two neighboring extreme  points is chosen as boundary of the object. Area within the boundaries is considered as foreground and the region outside the extreme points is considered as background for the appearance model. Grabcut (\cite{rother2004grabcut}) and Graphcut (\cite{kwatra2003graphcut}) are classic interactive segmentation models, which segment objects by gradually updating the appearance model. These models require the user to mark in both background and foreground regions. Although they use  extensive guiding signals, they would fail if the object has blurred or complex boundaries.
  
In recent years, CNN models have been extensively used for interactive segmentation (\cite{xu2017deep, xu2016deep, agustsson2019interactive, papadopoulos2017extreme, maninis2018deep, ling2019fast, castrejon2017annotating, acuna2018efficient, wang2019object}). A well-known example is DEXTRE (\cite{maninis2018deep}) which utilizes extreme points as an auxiliary input to the network. First, the annotator clicks four points on the extreme positions of objects then a heat map (Gaussian map for each point where points are at the centers of Gaussians) channel is created form these clicks which is attached to the input and serves as guiding signal.

There are methods in the literature that require the user to draw a bounding box around the desired object. \cite{wang2018interactive} proposed a method for interactive medical images segmentation where an object of interest is selected by drawing a bounding box around it. Then a deep network is applied on a cropped image to obtain segmentation. They also have a refinement step based on Grabcut that takes squiggles from the user to highlight the foreground and background regions. This model is applicable for single object (an organ) segmentation in CT/MRI images where this organ has similar appearance and shape in all images. However, this approach is not practical for segmentation of multiple objects (like nuclei) or amorphous objects (like glands) in histology domain. Some methods combined bounding box annotations with Graph Convolutional Network (GCN) to achieve interactive segmentation (\cite{ling2019fast,castrejon2017annotating, acuna2018efficient}). In these methods the selected bounding box is cropped from the image and fed to a GCN to predict polygon/spline around object. The polygon surrounds the object then can be adjusted in an iterative manner by refining the deep model.
Also, there are some hybrid methods which are based on the level sets (\cite{caselles1997geodesic}). \cite{acuna2019devil} and \cite{wang2019object} embedded the level set optimization strategy  in deep network to achieve precise boundary prediction from coarse annotations.

For some objects such as nuclei, manual selection of four extreme points or drawing a bounding box is still time-consuming, considering that an image of size 512$\times$512 can contain more that 200 nuclei. Moreover, extreme points for objects like glands are not providing sufficient guidance to delineate boundaries due to  complex shape and unclear edges of such objects. In this paper, we propose to use a single click or a squiggle as the guiding signal to keep simplicity in user interactions while providing enough information.
Similar to our approach is a work by \cite{sakinis2019interactive},  where the annotator needs to place two pairs of click points inside and outside of the object of interest. However, their method is limited to segmenting a single predefined object, like prostate organ in CT images
unlike the multiple objects (nuclei, cell, and glands) in histology images, as is the case in this study, that mutate greatly in appearance for different cases, organs, sampling/staining methods, and diseases.

\subsection{Interactive full image segmentation}
 Several methods have been proposed to interactively segment all objects within the visual field.  \cite{andriluka2018fluid} introduced Fluid Annotation, an intuitive human-machine interface for annotating the class label and delineating every object and background region in an image. An interactive version of Mask-RCNN (\cite{he2017mask}) was proposed by \cite{agustsson2019interactive} which accepts bounding box annotations and incorporates  a pixel-wise loss  allowing regions to compete on the common image canvas. Other older works that also segment full image are proposed by \cite{nieuwenhuis2012spatially, nieuwenhuis2014co, santner2010interactive, vezhnevets2005growcut}. 
 
Our method is different from these approaches as these are designed to segment all objects in natural scenes, requiring the user to label the background region and missing instances may interfere with the segmentation of  desired objects. Besides, these approaches require high degree of user interaction for each object instance (minimum of selecting 4 extreme points). However, in interactive segmentation of nuclei/cells from microscopy images, selecting four points for each object is very cumbersome. On the other hand, all above-mentioned methods are sensitive to the correct selection of extreme points which also can be very confusing for the user when he/she aims to mark a cancerous gland in histology image with complex shape and vague boundaries. Furthermore, another problem with a full image segmentation method like \cite{agustsson2019interactive} is that it uses Mask-RCNN backbone for RoI feature extraction which has difficulty in detecting objects with small sizes such as nuclei.
 
In this paper we propose \textbf{NuClick} that uses only one point for delineating nuclei and cells and a squiggle  for outlining glands. For nucleus and cell segmentation, proving a dot inside nucleus and cell is fast, easy, and does not require much effort from user compared to recent methods which rely on bounding boxes around objects. For glands, drawing a squiggle inside the glands is not only much easier and user friendly for annotator but also gives more precise annotations compared to other methods.
Our method is suitable for single object to full image segmentation and is applicable to a wide range of object scales, i.e. small nuclei to large glands. To avoid interference of neighboring objects in segmentation of desired object, a hybrid weighted loss function is incorporated in NuClick training.

This paper is complementary to our previous paper (\cite{jahanifar2019nuclick}), where we showed results of the preliminary version of NuClick and its application to nuclei, whereas here we extend its application to glands and cells. As a result of the current framework, we release two datasets of lymphocyte segmentation in Immunohistochemistry (IHC) images and segmentation mask of white blood cells (WBC) in blood sample images\footnote{https://github.com/navidstuv/NuClick}.

A summary of our contributions is as follows:
\begin{itemize}
    \item We propose the first interactive deep learning framework to facilitate and speed up collecting reproducible and reliable annotation in the field of computational pathology.
    \item We propose a deep network model using guiding signals and multi-scale blocks for precise segmentation of microscopic objects in a range of scales.
    \item We propose a method based on morphological skeleton for extracting guiding signals from gland masks, capable of identifying holes in objects.
    \item We Incorporate a weighted hybrid loss function in the training process which helps to avoid interference of neighboring objects when segmenting the desired object.
    \item Performing various experiments to show the effectiveness and generalizability of the NuClick.
    \item We release two datasets of lymphocyte dense annotations in IHC images and touching white blood cells (WBCs) in blood sample images.
\end{itemize}

\section{Methodology}
\subsection{NuClick framework overview}

Unlike previous methods that use a bounding box or at least four points \cite{maninis2018deep, boykov2001interactive, wu2014milcut, rother2012interactive, papadopoulos2017extreme} for interactive segmentation, in our proposed interactive segmentation framework only one click inside the desired object is sufficient. 
We will show that our framework is easily applicable for segmenting different objects in different levels of complexity.  We present a framework that is applicable for collecting segmentation for nuclei which are smallest visible objects in histology images, then cells which consist of nucleus and cytoplasm, and glands which are a group of cells.
Within the current framework the minimum human interaction is utilized to segments desired object with high accuracy. The user input for nucleus and cell segmentation is as small as one click and for glands a simple squiggle would suffice.

NuClick is a supervised framework based on convolutional neural networks which uses an encoder-decoder network architecture design.
In the training phase, image patches and guiding signals are fed into the network, therefore it can learn where to delineate objects when an specific guiding signal appears in the input. In the test phase, based on the user-input annotations (clicks or squiggles), image patches and guiding signal maps are generated to be fed into the network. Outputs of all patches are then gathered in a post-processing step to make the final instance segmentation map.
We will explain in details all aspects of this framework in the following subsections.

\subsection{Model architecture \& loss}
\begin{figure*} [ht!]
\includegraphics[width =\textwidth]{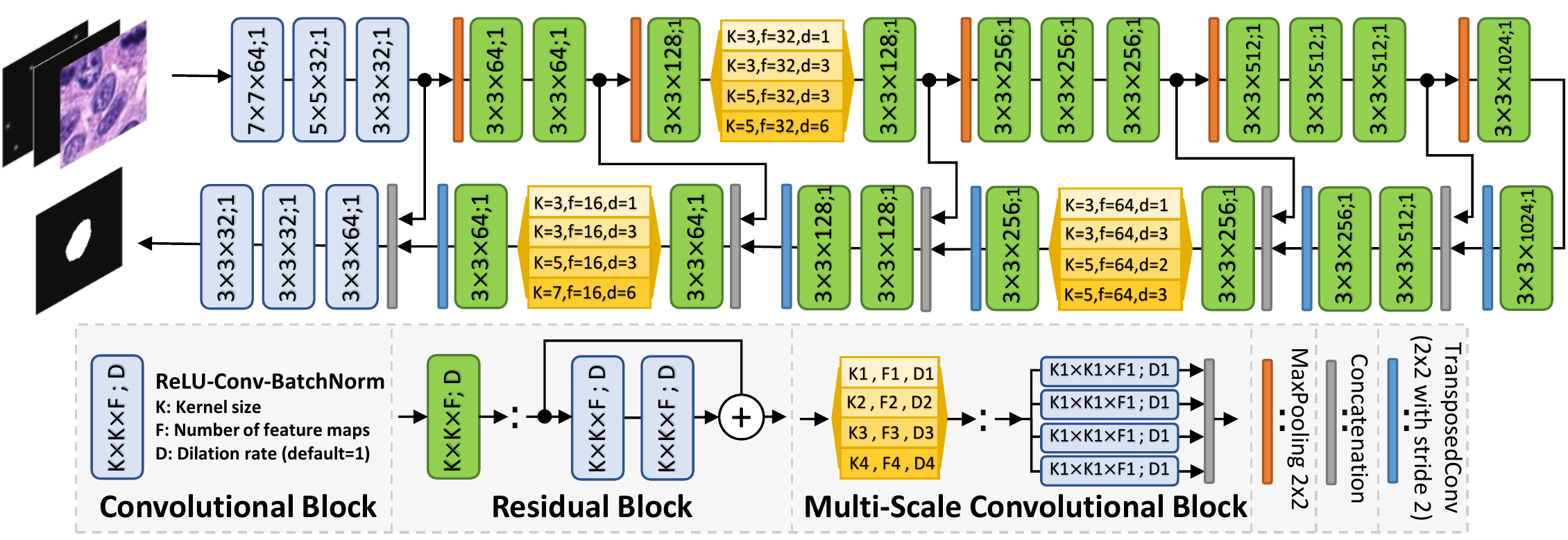}
\caption{Overview of the NuClick network architecture which consists of Convolutional, Residual, and Multi-Scale convolutional blocks.}
\label{fig:network architecture}
\end{figure*}

Efficiency of using encoder-decoder design paradigm for segmentation models has been extensively investigated in the literature and it has been shown that UNet  design paradigm works the best for various medical (natural) image segmentation tasks (\cite{hesamian2019deep, garcia2017review}). Therefore, similar to \cite{jahanifar2019nuclick}, an encoder-decoder architecture with multi-scale and residual blocks has been used for NuClick models, as depicted in \cref{fig:network architecture}.

As our goal is to propose a unified network architecture that segments various objects (nuclei, cells and glands), it must be capable of recognizing objects with different scales. In order to segment both small and large objects, the network must be able to capture features on various scales. Therefore, we incorporate multi-scale convolutional blocks \cite{jahanifar2018segmentation} throughout the network (with specific design configurations related to the network level). Unlike other network designs (eg. DeepLab v3 \cite{chen2017rethinking}) that only uses multi-scale \textit{atrous} convolutions in the last low-resolution layer of the encoding path, we use them in three different levels both in encoding and decoding paths. By doing this, NuClick network is able to extract relatable semantic multi-scale features from the low-resolution feature maps and generate fine segmentation by extending the receptive fields of its convolution layers in high-resolution feature maps in the decoder part. Parameters configuration for residual and multi-scale blocks is shown on each item in the \cref{fig:network architecture}

Furthermore, using residual blocks instead of plain convolutional layers enables us to design a deeper network without risk of gradient vanishing effect \cite{he2016deep}.
In comparison to \cite{jahanifar2019nuclick}, the network depth has been further increased to better deal with more complex objects like glands.

The loss function used to train NuClick is a combination of soft dice loss and weighted cross entropy. The dice loss helps to control the class imbalance and the weighted cross entropy part penalizes the loss if in the prediction map other objects rather than the desired object were present.
\begin{equation}
\begin{array}{l}
{\cal L} = 1 - {{\left( {\sum\limits_i {{p_i}{g_i} + \varepsilon } } \right)} \mathord{\left/
 {\vphantom {{\left( {\sum\limits_i {{p_i}{g_i} + \varepsilon } } \right)} {\left( {\sum\limits_i {{p_i} + \sum\limits_i {{g_i} + \varepsilon } } } \right)}}} \right.
 \kern-\nulldelimiterspace} {\left( {\sum\limits_i {{p_i} + \sum\limits_i {{g_i} + \varepsilon } } } \right)}}\\
\quad \;\; - \frac{1}{n}\sum\limits_{i = 1}^n {{w_i}({g_i}\log {p_i} + (1 - {g_i})\log (1 - {p_i}))} 
\end{array}
\label{eq:loss}
\end{equation}
where $n$ is the number of pixels in the image spatial domain, ${p_i}$, ${g_i}$, and ${w_i}$ are values of the prediction map, the ground-truths mask $\bf{G}$, and the weight map $\bf{W}$ at pixel $i$, respectively and $\varepsilon$ is a small number. Considering that $\bf{G}$ has value of 1 for the desired (included) objects and 0 otherwise, its complement ${\bf{\tilde G}}$ has value of 1 for the undesired (excluded) objects in the image and 0 otherwise. The adaptive weight map is then defined as: ${\bf{W}} = {\alpha}^2 {\bf{G}} + \alpha{\bf{\tilde G}}+1$ ,where $\alpha$ is the adaptive factor that is defined based on areas of the included and excluded objects as follows: $
\alpha  = \max \left\{ {{{\sum {{\bf{\tilde G}}} } \mathord{\left/
 {\vphantom {{\sum {{\bf{\tilde G}}} } {\sum {\bf{G}} }}} \right.
 \kern-\nulldelimiterspace} {\sum {\bf{G}} }},1} \right\}$.
This weighting scheme puts more emphasis on the object to make sure it would be completely segmented by the network while avoiding false segmentation of touching undesired objects.

\subsection{Guiding Signals}

\subsubsection{Guiding signal for nuclei/cells}
When annotator clicks inside a nucleus, a map to guide the segmentation is created, where the clicked position is set to  one and the rest of pixels are set to zero which we call it \textit{inclusion map}.
In most scenarios, when more than one nucleus are clicked by the annotator (if he/she wants to have all nuclei annotated), another map is also created where positions of all nuclei except the desired nucleus/cell are set to one and the rest of pixels are set to zero, which is called \textit{exclusion map}. When only one nucleus is clicked exclusion map is a zero map.
Inclusion and exclusion maps are concatenated to RGB images to have 5 channels as the input to the network (as illustrated in \cref{fig:network architecture}).
The same procedure is used for creating guiding signals of cells.
However, we took some considerations into the training phase of the NuClick in order to make it robust against guiding signal variations. In the following paragraphs, we will describe these techniques for both training and testing phases.
\paragraph{Training}
To construct inclusion map for training, a point inside a nucleus/cell is randomly chosen. It has been taking into account that the sampled point has at least 2 pixels distance from the object boundaries. 
The exclusion map on the other hand is generated based on the centroid location of the rest of nuclei within the patch. 
Thereby, guiding signals for each patch are continuously changing during the training. Therefore the network sees variations of guiding signals in the input for each specific nuclei and will be more robust against human errors during the test. In other words the network learns to work with click points anywhere inside the desired nuclei so there is no need of clicking in the exact centroid position of the nuclei.

\paragraph{Test}
At inference time, guiding signals are simply generated based on the clicked positions by the user. For each desired click point on image patch, an inclusion map
and an exclusion map
are generated.
The exclusion map have values if user clicks on more than one nuclei/cells, otherwise it is zero. Size of information maps for nuclei and cells segmentation tasks are set to $128\times128$ and $256\times256$, respectively. For test time augmentations we can disturb the position of clicked points by 2 pixels in random direction.
The importance of exclusion map is in cluttered areas where nuclei are packed together. If the user clicks on all nuclei within these areas, instances will be separated clearly. In the experimental section we will show the effect of using exclusion maps.

\subsubsection{Guiding signal for glands}
Unlike nuclei or cells, since glands are larger and more complex objects, single point does not provide strong supervisory signal to the network. Therefore, we should chose another type of guiding signal which is informative enough to guide the network and simple enough for annotator during inference. Instead of points, we propose to use squiggles. More precisely, the user provides a squiggle inside the desired gland which determines the extent and connectivity of it.

\paragraph{Training}
Considering $\bf{M}$ as the desired ground truth (GT) mask in the output, an inclusion signal map is randomly generated as follows: First we apply a Euclidean distance transform function $D(x)$ on the mask to obtain distances of each pixel inside the mask to the closest point on the object boundaries:
\begin{equation}
    {D_{i,j}}(\bf{M}) = \left\{ {\sqrt {{{(i - {i_b})}^2} + {{(j - {j_b})}^2}} |(i,j) \in \bf{M}} \right\}
\end{equation}
where ${i_b}$ and ${j_b}$ are the closest pixel position on the object boundary to the desired pixel position $(i,j)$.
Afterwards, we select a random threshold ($\tau$) to apply on the distance map for generating a new mask of the object which indicates a region inside the original mask.
$$
{{\overline \bf{M}}_{i,j}} = \left\{ \begin{array}{l}
1\,\,\,\,\,\,\,if\,\,{D_{i,j}} > \tau \\
0\,\,\,\,\,\,otherwise\,
\end{array} \right.
$$

The threshold is chosen based on the mean ($\mu$) and standard deviation ($\sigma$) of outputs of distance function, where the interval for choosing $\tau$ is $[0 ,\mu  + \sigma ]$.

Finally, to obtain the proper guiding signal for glands, the morphological skeleton (\cite{serra1983image}) of the new mask ${\overline \bf{M}}$ is constructed. Note that we could have used the morphological skeleton of the original mask as the guiding signal (which does not change throughout the training phase) but that may cause the network to overfit towards learning specific shapes of skeleton and prevents it from adjusting well with annotator input. Therefore, by changing the shape of the mask, we change the guiding signal map during training.
\begin{figure*}[ht]
\centering
\includegraphics[width =0.9\textwidth]{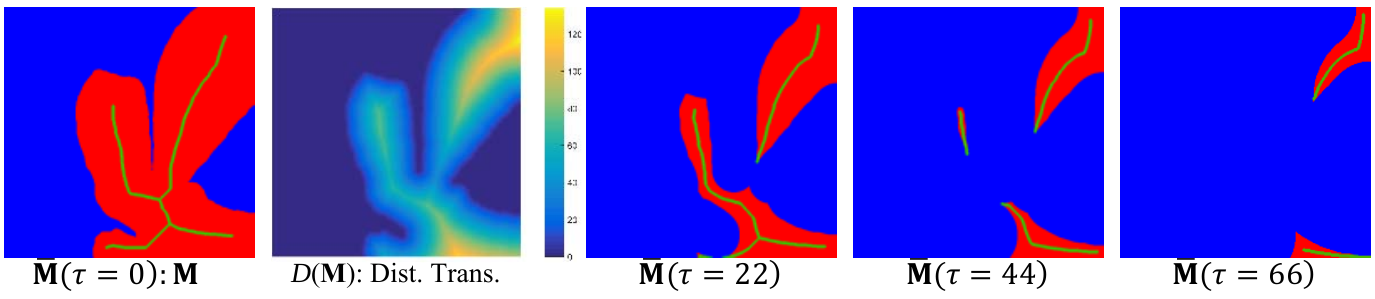}
\caption{Generating supervisory signal (inclusion map) for NuClick while training on gland dataset. The left image is the GT mask of a sample gland and $D(\bf{M})$ is the distance transformation of that mask. By changing the threshold value ($\tau$), the guiding signal (skeleton of the new mask ${\overline \bf{M}}$ which is specified by green color) is also changing.}
\label{fig:glandGuid}
\end{figure*}
An example of constructing map for a gland is depicted in the \cref{fig:glandGuid}. In this figure, the left hand side image represents the GT of the desired gland on which its corresponding skeleton is overlaid with green color. If we use this same mask for training the network, the guiding signal would remain the exact same for all training epochs. However, based on our proposed mask changing technique, we first calculate the distance transformation of the GT, $D(\bf{M})$, and then apply a threshold of $\tau$ on it to construct a new mask of  ${\overline \bf{M}}$. As you can see in Fig. \ref{fig:glandGuid}, by changing the the threshold value, appearance of the new mask is changing which results in different morphological skeletons as well (note the change of overlaid green colored lines with different $\tau$ values). This will make the NuClick network robust against the huge variation of guiding signals provided by the user during the test phase.
The exclusion map for gland is constructed similar to nuclei/cells i.e., except one pixel from each excluding object all other pixels are set to zero.


\paragraph{Test}
When running inference, the user can draw squiggles inside the glandular objects. Then patches of 512$\times$512 are extracted from image based on the bounding box of the squiggle. If the bounding box height or width is smaller than 512, it is relaxed until height and width are 512. And if the bounding box is larger than 512 then image and corresponding squiggle maps are down-scaled to 512$\times$512. 

\subsection{Post-processing}
After marking the desired objects by the user, image patches, inclusion and exclusion maps are generated and fed into the network to predict an output segmentation for each patch. Location of each patch is stored in the first step, so it can be used later to build the final instance segmentation map.

 The first step in post-processing is converting the prediction map into an initial segmentation mask by applying a threshold of 0.5. Then small objects (objects with area less than 50 pixels) are removed. Moreover, for removing extra objects except desired nucleus/cell/gland inside the mask, morphological reconstruction operator is used. To do so, the inclusion map plays the role of marker and initial segmentation is considered as the mask in morphological reconstruction.

\section{Setups and Validation Experiments}
\subsection{Datasets}
\paragraph{Gland datasets} Gland Segmentation dataset \cite{sirinukunwattana2017gland} (GlaS) and GRAG datasets \cite{awan2017glandular, graham2019mild} are used for gland segmentation. GlaS dataset consists of 165 tiles, 85 of which for training and  80 for test. Test images of GlaS dataset are also split into to TestA and TestB. TestA was released to the participants of the GlaS challenge one
month before the submission deadline, whereas Test B was released on the final day of the challenge. Within GRAG dataset, there are a total of 213 images which is split into  173
training images and 40 test images with different cancer grades. Both of these datasets are extracted from Hematoxylin and Eosin (H\&E) WSIs.

\paragraph{Nuclei dataset} MonuSeg (\cite{kumar2019multi}) and CPM (\cite{vu2019methods}) datasets which contain 30 and 32 H\&E images ,respectively, have been used for our experiments. 16 images of each of these datasets are used for training.
\paragraph{Cell dataset} A dataset of 2689 images consisting of touching white blood cells (WBCs) were synthetically generated for cell segmentation experiments. To this end, we used a set of 11000 manually segmented non-touching WBCs (WBC library). Selected cells are from one of the main five category of WBCs: Neutrophils, Lymphocytes, Eosinophils, Monocytes, or Basophils.

The original patches of WBCs were extracted from scans of peripheral blood samples captured by CELLNAMA LSO5 slide scanner equipped with oil immersion 100x objective lens. However, the synthesized images are designed to mimic the appearance of bone marrow samples. In other words, synthesized images should contain several (10 to 30) touching WBCs. Therefore, for generating each image a random number of cells are selected from different categories of WBC library and then they are added to a microscopic image canvas which contains only red blood cells. During the image generation each added cell is well blended into the image so its boundary looks seamless and natural. This would make the problem of touching object segmentation as hard as real images. It is worth mentioning that each WBC is augmented (deformed, resize, and rotate) before being added to the canvas. Having more than 11000 WBCs and performing cell augmentation during the image generation would guarantee that the network does not overfit on a specific WBC shape.
For all datasets 20\% of training images are considered as validation set.

\subsection{Implementation Details}
For our experiments, we used a work station equipped with an Intel Core i9 CPU, 128GB of RAM and two GeForce GTX 1080 Ti GPUs. All experiments were done in Keras framework with Tensorflow backend. For all applications, NuClick is trained for 200 epochs. Adam optimizer with learning rate of $3 \times {10^{ - 3}}$ and weight decay of of $5 \times {10^{ - 5}}$ was used to train the models. Batch size for nuclei, cell and gland was set to 256, 64 and 16 respectively. We used multiple augmentations as follows: random horizontal and vertical flip, brightness adjustment, contrast adjustment, sharpness adjustment, hue/saturation adjustment, color channels shuffling and adding Gaussian noise (\cite{jahanifar2018segmentation}). 

\subsection{Metrics}
For our validation study we use metrics that has been reported in the literature for cell and gland instance segmentation. For nuclei and cells we have used AJI (Aggregated Jaccard Index) proposed by \cite{kumar2017dataset}: an instance based metric which calculates Jaccard index for each instance and then aggregates them, Dice coefficient: A similar metric to IoU (Intersection over Union), Hausdorff distance (\cite{sirinukunwattana2017gland}): the distance between two polygons which is calculated per object, Detection Quality (DQ): is equivalent to ${F\textsubscript{1}}-Score$ divided by 2, SQ: is summing up IoUs for all true positive values over number of true positives and PQ: DQ$\times$SQ (\cite{kirillov2019panoptic}). For AJI, Dice, the true and false values are based on the pixel value but for DQ true and false values are based on the value of IoU. The prediction is considered true positive if IoU is higher 0.5.\\
\begin{table}
\centering
\caption{Comparison of the proposed network architecture with other approaches: MonuSeg dataset have been used for these experiments.}
\label{tArch}
\begin{tabular}{lllll} 
\hline\hline
                      & AJI             & Dice            & PQ              & Haus.           \\ 
\hline
Unet                  & 0.762           & 0.821           & 0.774           & 8.73            \\
FCN                   & 0.741           & 0.798           & 0.756           & 9.5             \\
Segnet                & 0.785           & 0.846           & 0.794           & 8.33            \\
NuClick W/O MS block  & 0.798           & 0.860           & 0.808           & 6.11            \\
NuClick + 1 MS block  & 0.817           & 0.889           & 0.820           & 5.51            \\
NuClick + 2 MS blocks & 0.830           & 0.905           & 0.829           & 4.93            \\
NuClick + 3 MS blocks & 0.834           & 0.912           & \textbf{0.838}  & \textbf{4.05}   \\
NuClick + 4 MS blocks & \textbf{0.835}  & \textbf{0.914}  & 0.838           & 4.05            \\
\hline\hline
\end{tabular}
\end{table}
For gland segmentation, we use F1-score, Dice\textsubscript{Obj}, and Hausdorff distance (\cite{sirinukunwattana2017gland}). The True positives in F1-score are based on the thresholded IoU. Dice\textsubscript{Obj} is average of dice values over all objects and Hausdorff distance here is the same as the one used for nuclei.

\subsection{Network Selection}
In this section, we investigate the effect of  multi-scale blocks on NuClick network and compare its performance with other popular architectures. Ablating various choices of components in NuClick network architecture have been shown in \cref{tArch}. We tested our architecture with up to 4 multi-scale (MS) blocks and we observed that adding more that 3 MS blocks does not contribute significantly to the performance. It can be observed that our architecture outperforms three other popular methods (UNet  by \cite{ronneberger2015u}, SegNet by \cite{badrinarayanan2017segnet},  and FCN by \cite{long2015fully}). When we use no MS block, our model is still better than all baseline models which shows the positive effect of using residual blocks. We opt to use 3 MS blocks in the final NuClick architecture because it is suggesting a competitive performance while having smaller network size.

\subsection{Validation Experiments}

Performance of NuClick framework for interactive segmentation of nuclei, cells, and glands are reported in \cref{tValNuc,tValCell,tValGland}, respectively. For nuclei and cells, centroid of the GT masks were used to create inclusion and exclusion maps, whereas for gland segmentation, morphological skeleton of the GT masks were utilized.
For comparison purposes, performance of other supervised and unsupervised interactive segmentation methods are included as well. In \cref{tValNuc,tValCell}, reported methods are Region Growing (\cite{adams1994seeded}): iteratively determines if the neighbouring pixels of an initial seed point should belong to the initial region or not (in this experiment, the seed point is GT mask centroid and the process for each nuclei/cell is repeated 30 iterations),  Active Contour (\cite{chan2001active}): which iteratively evolves the level set of an initial region based on internal and external forces (the initial contour in this experiment is a circle with radius 3 pixels positioned at the GT mask centroid),  marker controlled watershed (\cite{parvati2008image}) that is based on watershed algorithm in which number and segmentation output depends on initial seed points (in this experiment, unlike \cite{parvati2008image} that generates seed points automatically, we used GT mask centroids as seed points), interactive Fully Convolutional Network--iFCN (\cite{xu2016deep}): a supervised DL based method that transfers user clicks into distance maps that are concatenated to RGB channels to be fed into a fully convolutional neural network (FCN), and Latent Diversity--LD (\cite{li2018interactive}): which uses two CNNs to generate final segmentation. The first model takes the image and distance transform of two dots (inside and outside of object) to generate several diverse initial segmentation maps and the second model selects the best segmentation among them.

\begin{table}
\centering
\caption{Performance of different interactive segmentation methods for nuclear segmentation on validation set of MonuSeg dataset}
\label{tValNuc}
\begin{tabular}{lccccl} 
\hline\hline
Method         & AJI                       & Dice                      & SQ                        & PQ                        & Haus.  \\ 
\hline
Watershed      & 0.189                     & 0.402                     & 0.694                     & 0.280                     & 125    \\
Region Growing & 0.162                     & 0.373                     & 0.659                     & 0.241                     & 95    \\
Active Contour & 0.284                     & 0.581                     & 0.742                     & 0.394                     & 67    \\
iFCN           & 0.806                     & 0.878                     & 0.798                     & 0.782                     & 7.6    \\
LD          & 0.821                     & 0.898                     & 0.815                     & 0.807                     & 5.8    \\

NuClick        & \textbf{0.834  }                   & \textbf{0.912}                     & \textbf{0.839}                     & \textbf{0.838}                  & \textbf{4.05}   \\
\hline\hline
\end{tabular}
\end{table}


\begin{table}
\centering
\caption{Performance of different interactive segmentation methods for cell segmentation on test set of WBC dataset}
\label{tValCell}
\begin{tabular}{lccccl} 
\hline\hline
               & AJI                        & Dice                      & SQ                        & PQ                        & Haus.  \\ 
\hline
Watershed      & 0.153                      & 0.351                     & 0.431                     & 0.148                     & 86     \\
Region Growing & 0.145                      & 0.322                     & 0.414                     & 0.129                     & 71     \\
Active Contour & 0.219                      & 0.491                     & 0.522                     & 0.198                     & 50     \\
iFCN           & 0.938                      & 0.971                     & 0.944                     & 0.944                     & 9.51   \\
LD           & 0.943                     & 0.978                     & 0.949                     & 0.949                     & 8.33    \\
NuClick        & \textbf{0.954}                      & \textbf{0.983}                     & \textbf{0.958}                     & \textbf{0.958}                    & \textbf{7.45}   \\
\hline\hline
\end{tabular}
\end{table}

\begin{table}
\centering
\setlength{\tabcolsep}{2.5pt}
\caption{Performance of different interactive segmentation methods for gland segmentation on test sets of GLaS dataset}
\label{tValGland}
\begin{tabular}{lccc|ccc} 
\hline\hline
            & \multicolumn{3}{c|}{TestA}                  & \multicolumn{3}{c}{TestB}                    \\ 
\hline
            & F1                        & Dice\textsubscript{Obj} & Haus. & F1                        & Dice\textsubscript{Obj} & Haus.  \\ 
\hline
Grabcut     & 0.462                     & 0.431   & 290   & 0.447                     & 0.412   & 312    \\
Deep Gabcut & 0.886                     & 0.827   & 51    & 0.853                     & 0.810   & 57     \\
DEXTRE      & 0.911                     & 0.841   & 43    & 0.904                     & 0.829   & 49     \\
Mask-RCNN  & \multicolumn{1}{l}{0.944} & 0.875   & 35    & \multicolumn{1}{l}{0.919} & 0.856   & 41     \\
BIFseg  & \multicolumn{1}{l}{0.958} & 0.889   & 28    & \multicolumn{1}{l}{0.921} & 0.864   & 38     \\
NuClick     & \textbf{1.000}                     & \textbf{0.956}   & \textbf{15}    & \textbf{1.000}                      & \textbf{0.951}   & \textbf{21}     \\
\hline\hline
\end{tabular}
\end{table}

In \cref{tValGland}, reported methods are Grabcut by \cite{rother2004grabcut}:  which updates appearance model within the bounding box provided by the user,  Deep GrabCut by \cite{xu2017deep}: which converts the bounding box provided by the user into  a distance map  that is concatenated to RGB image as the input of a deep learning model, DEXTRE (\cite{maninis2018deep}): a supervised deep learning based method  which is mentioned in the \cref{Sec-inter} and accepts four extreme points of glands as input (extreme points are extracted based on each object GT mask), and a Mask-RCNN  based approach proposed by \cite{agustsson2019interactive}: where the bounding box is also used as the input to the Mask-RCNN. \cite{agustsson2019interactive} also added a instance-aware loss measured at the pixel level to the Mask-RCNN loss.  We also compared our method for gland segmentation with BIFseg (\cite{wang2018interactive}) that needs user to crop the object of interest by drawing bounding box around it. The cropped region is then resized and fed into a resolution-preserving CNN to predict the output segmentation. \cite{wang2018interactive} also used a refinement step which is not included in our implementation.

For GrabCut, Deep GrabCut, BIFseg, and Mask-RCNN approaches the bounding box for each object is selected based on its GT mask. For iFCN and LD methods, positive point (point inside the object) is selected according to the centroid of each nucleus and negative click is a random point outside the desired object.

\begin{figure*}[h!]
\centering
\includegraphics[width =0.95\textwidth]{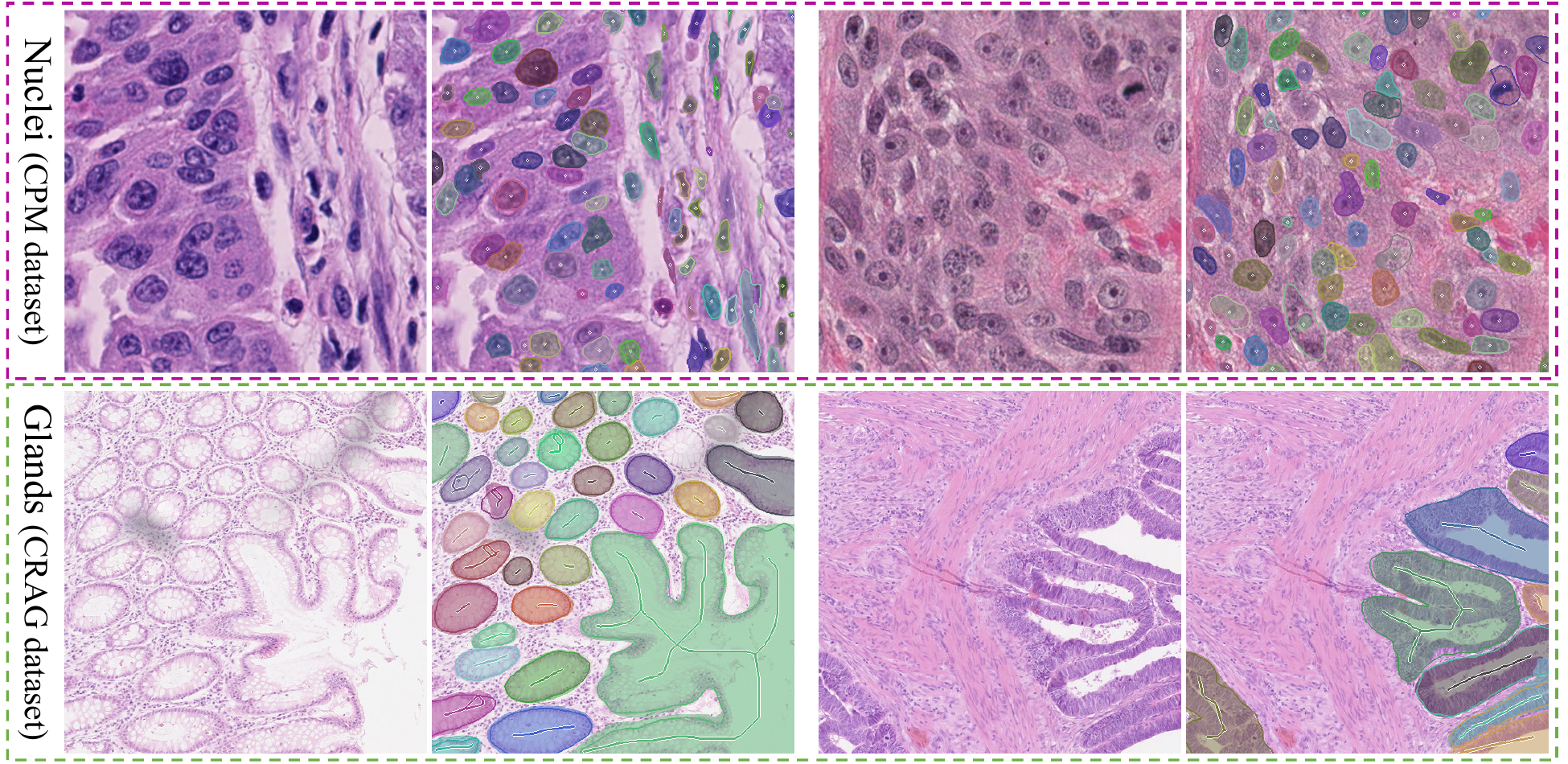}
\caption{\textbf{Generalizability of NuClick:} The first row shows results of NuClick on CPM dataset for nuclei segmentation (where the network was trained on MoNuSeg dataset). The second row illustrates two samples of gland segmentation task from CRAG dataset  where the model was trained on GLaS dataset. Solid  stroke  line  around  each  object  outlines the  ground  truth  boundary  for  that  object,  overlaid  transparent  mask  is  the predicted segmentation region by NuClick, and points or squiggles indicate the provided guiding signal for interactive segmentation. (Best viewed in color)}
\label{fig:generalize}
\end{figure*}

Based on \cref{tValNuc}, NuClick achieved AJI score of 0.834, Dice value of 0.912, and PQ value of 0.838 which outperformed all other methods for nuclear segmentation on MonuSeg dataset. Performance gap between NuClick and other unsupervised methods is very high (for example in comparison with Watershed method, NuClick achieves a 0.645 higher AJI). Extreme low evaluation values achieved by unsupervised metrics indicate that they are not suitable for intricate task of nuclear segmentation, even if they are fed with GT markers.
There is also iFCN (\cite{xu2016deep}), a deep learning based method in \cref{tValNuc} that is trained based on the clicked dots inside and outside of objects. However, NuClick performs better than iFCN for all AJI, Dice, and PQ metrics by margin of 2.8\%, 3.4\%, and 5.6\%, respectively, which is a considerable boost. For the other CNN based method in \cref{tValNuc}, LD method, NuClick advantage over all metrics is also evident.

The same performance trend can be seen for both cell and gland segmentation tasks in \cref{tValCell,tValGland}. For the cell segmentation task, NuClick was able to segment touching WBCs from synthesized dense blood smear images quite perfectly. Our proposed method achieves AJI, Dice, and PQ values of 0.954, 0.983, and 0.958, respectively, which indicates remarkable performance of the NuClick in cell segmentation. 

Validation results of our algorithm on two test sets from GlaS dataset (testA and testB) are reported in \cref{tValGland} alongside the results of 4 supervised deep learning based algorithms and an unsupervised method (Grabcut). Markers used for Grabcut are the same as ones that we used for NuClick. Based on \cref{tValGland} our proposed method is able to outperform all other methods for gland segmentation in both testA and testB datasets by a large margine. For testB, NuClick achieves F1-score of 1.0, Dice similarity coefficient of 0.951, and Hausdorff distance of 21, which compared to the best performing supervised method (BIFseg) shows 7.9\%, 8.7\%, and 17 pixels improvement, respectively. The F1-score value of 1.0 achieved for NuClick framework in gland segmentation experiment expresses that all of desired objects in all images are segmented well enough. As expected, unsupervised methods, like Grabcut, perform much worse in comparison to supervised method for gland segmentation. Quantitatively, our proposed framework shows  55.3\% and 53.9\% improvement compared to Grabcut in terms of F1-score and Dice similarity coefficients. The reason for the advantage of NuClick over other methods mainly lies in its squiggle-based guiding signal which is able to efficiently mark the extent of big, complex, and  hollow objects. It is further discussed in \cref{Sec:discussion}.

Methods like DEXTRE, BIFseg,  and Mask-RCNN are not evaluated for interactive nucleus/cell segmentation, because they may be cumbersome to apply in this case. These methods need four click points on the boundaries of nucleus/cell (or drawing a bounding box for each of them) which is still labour-intensive as there may be a large number of nuclei/cells within an image.

Segmentation quality for three samples are depicted in \cref{fig:validation}. In this figure, the first, second, and third rows belong to a sample drawn from MoNuSeg, WBC, and GLaS validation sets. The left column of \cref{fig:validation} shows original images and images on the right column contains GT boundaries, segmentation mask, and guiding signals (markers) overlaid on them. Guiding signals for nuclei and cell segmentation are simple clicks inside each object (indicated by diamond-shape points on the images) while for glands (the third row) guiding signals are squiggles. In all exemplars, extent of  the prediction masks (indicated by overlaid transparent colored region) are very close to the GT boundaries (indicated by solid strokes around each object).

\section{Discussions}
\label{Sec:discussion}
In order to gain better insights into the performance and capabilities of the NuClick, we designed several evaluation experiments. In this section we will discuss different evaluation experiments for NuClick. First we will assess the generalizability of the proposed framework, then we will discuss how it can adapt to new domains without further training, after that the reliability of NuClick output segmentation is studied. Moreover, sensitivity of output segmentation to variations in the guiding signals is also addressed in the following subsections.

 \begin{figure*}[ht!]
\centering
\includegraphics[width =0.95\textwidth]{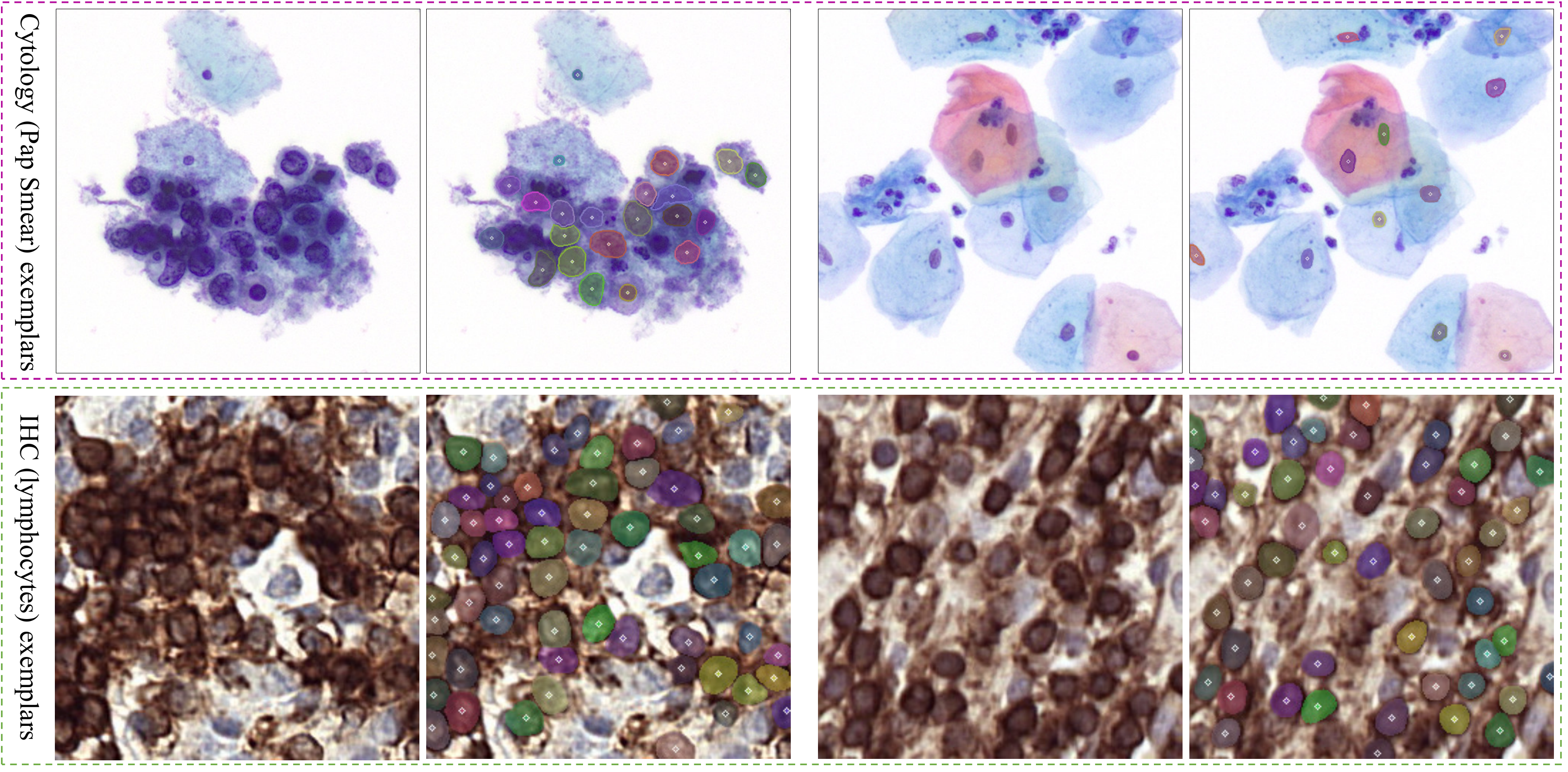}
\caption{\textbf{Domain adaptability of NuClick:} nuclei from unseen domains (Pap Smear sample in the first row and IHC stained sample in the second tow) are successfully segmented using the NuClick which was trained on MoNuSeg dataset. In all images, solid stroke line around each object outlines the ground truth boundary for that object (except for IHC samples, for which ground truth masks are unavailable),  overlaid transparent mask is the predicted segmentation region by NuClick, and points  indicate the provided guiding signal for interactive segmentation. (best viewed in color)}
\label{fig:adapt}
\end{figure*}

 \subsection{Generalization study}
 To show the generalizability of the NuClick across an unseen datasets, we designed an experiment in which NuClick is trained on the training set of a specific dataset and then evaluated on the validation set of another dataset but within the same domain. Availability of different labeled nuclei and gland datasets allow us to better show the generalizability of our proposed framework across different dataset and different tasks.

 To assess the generalizability upon nuclei segmentation, two experiments were done. In one experiment, NuClick was trained on training set of MoNuSeg dataset and then evaluated on the validation set of CPM dataset. In another experiment this process was done contrariwise where CPM training set was used for training the NuClick and MoNuSeg testing set was used for the evaluation. Evaluation results of this study are reported in the first two rows of \cref{tGeneralize}.
From this table we can conclude that NuClick can generalize well across datasets because it gains high values for evaluation metrics when predicting images from dataset that was not included in its training. For example, when NuClick is trained on the MoNuSeg training set, Dice and SQ evaluation metrics resulted for CPM validation set are 0.908 and 0.821, respectively, which are very close to the values reported for evaluating the MoNuSeg validation set using the same model i.e., Dice of 0.912 and SQ of 0.839 in \cref{tValNuc}. This closeness for two different datasets using the same model supports our claim about generalizability of the NuClick.

Similarly, to test the generlizability of the NuClick when working on gland segmentation task, it has been trained on one gland dataset and tested on validation images from another gland dataset. As GlaS test set is divided into TestA and TestB, when NuClick is trained on CRAG, it has been test on testA and testB of GlaS (named as GlaSA and GlaSB in \cref{tGeneralize}). High values of Dice\textsubscript{Obj} metric and low values for Hasdroff distances also supports the generalizability of NuClick framework for gland segmentation task as well.

To provide visual evidence for this claim, we illustrated two nuclear segmentation samples from CPM validation set (resulted using a model trained on MoNuSeg dataset) and two gland segmentation samples from CRAG validation set (resulted using a model trained on GLaS dataset) in  \cref{fig:generalize}. In all cases NuClick was able to successfully segment the desired objects with high accuracy. In all images of \cref{fig:generalize} different overlaid colors corresponds to different object instances, solid stroke lines indicate GT boundaries, transparent color masks show the predicted segmentation region, and other point or squiggle markers representing guiding signals for interactive segmentation.

\begin{table}
\centering
\setlength{\tabcolsep}{2pt}
\caption{Results of generalization study across different datasets for interactive nuclei and gland segmentation}
\begin{tabular}{lllcc|cc}
\hline\hline
                        & Train   & Test    & Dice                  & SQ                     & Dice\textsubscript{Obj}                 & Haus.                   \\ 
\hline
\multirow{2}{*}{Nuclei} & MoNuSeg & CPM     & 0.908                  & 0.821                   & -                        & -                       \\
                        & CPM     & MoNuSeg & 0.892                  & 0.811                   & -                        & -                       \\ 
\hdashline
\multirow{2}{*}{Gland}  & GLaS    & CRAG    & \multicolumn{1}{l}{-} & \multicolumn{1}{l|}{-} & \multicolumn{1}{l}{0.932} & \multicolumn{1}{l}{31}  \\
                        & CRAG    & GLaSA    & \multicolumn{1}{l}{-} & \multicolumn{1}{l|}{-} & \multicolumn{1}{l}{0.944} & \multicolumn{1}{l}{28} \\
                        & CRAG    & GLaSB    & \multicolumn{1}{l}{-} & \multicolumn{1}{l|}{-} & \multicolumn{1}{l}{0.938} & \multicolumn{1}{l}{30} \\
\hline\hline
\end{tabular}
\label{tGeneralize}
\end{table}
\subsection{Domain adaptation study}
To assess the performance of the NuClick on unseen samples from different data domains, we trained it on MoNuSeg  dataset which contains labeled nuclei from histopathological images and then used the trained model to segment nuclei in cytology and immunohistochemistry (IHC) samples.

In the cytology case, a dataset of 42 FoVs were captured from 10 different Pap Smear samples using CELLNAMA LSO5 slide scanner and 20x objective lens. These samples contain overlapping cervical cells, inflammatory cells, mucus, blood cells and debris. Our desired objects from these images are nuclei of cervical cells. All nuclei from cervical cells in the available dataset of Pap Smear images were manually segmented with the help of a cytotechnologist. Having the GT segmentation for nuclei, we can use their centroid to apply the NuClick on them (perform pseudo-interactive segmentation) and also evaluate the results quantitatively, as reported in \cref{tAdapt}. High values of evaluation metrics reported in \cref{tAdapt} shows how well NuClick can perform on images from a new unseen domain like Pap Smear samples. Some visual examples are also provided in fig. \ref{fig:adapt} to support this claim. As illustrated in the first row of fig. \ref{fig:adapt}, NuClick was able to segment touching nuclei (in very dense cervical cell groups) from Pap Smear samples with high precision. It is able to handle nuclei with different sizes and various background appearances. 

\begin{table}
\centering
\caption{Performance NuClick framework on segmenting nuclei in images from an unseen domain (Pap Smear)}
\begin{tabular}{lccccc} 
\hline\hline
Method         & AJI   & Dice  & SQ    & DQ    & PQ     \\ 
\hline
NuClick       & 0.934 & 0.965  & 0.933  & 0.997  & 0.931 \\ 
\hline\hline
\end{tabular}
\label{tAdapt}
\end{table}

For the IHC images, we utilized NuClick to delineate lymphocytes. The dataset we have used for this section is a set of 441 patches with size of $256\times256$ extracted from LYON19 dataset. LYON19 is scientific challenge on lymphocyte detection from images of IHC samples. In this dataset samples are taken from breast, colon or prostate organs and are then stained with an antibody against CD3 or CD8 \cite{lyon19} (membrane of lymphocyte would appear brownish in the resulting staining). However, for LYON19 challenge organizers did not release any instance segmentation/detection GTs alongside the image ROIs. Therefore, we can not assess the performance of NuClick segmentation on this dataset quantitatively. However, the quality of segmentation is very desirable based on the depicted results for two random cases in the second row of \cref{fig:adapt}. Example augmentations in  \cref{fig:adapt} are achieved by clicks of a non-expert user inside lymphocytes (based on his imperfect assumptions). As it is shown in \cref{fig:adapt}, NuClick is able to adequately segment touching nuclei even in extremely cluttered areas of images from an unseen domain. These resulting instance masks were actually used to train an automatic nuclei instance segmentation network, SpaNet \cite{koohbanani2019nuclear}, which helped us achieve the first rank in LYON19 challenge. In other words, we approached the problem lymphocyte detection as an instance segmentation problem by taking advantage of our own generated nuclei instance segmentation masks \cite{jahanifar2019nuclick}. 
It also approves the reliability of the NuClick generated prediction masks, which is discussed in more details in the following subsection.

\subsection{Segmentation Reliability Study}
The important part of an interactive method for collecting segmentation is to see how the generated segmentation maps are reliable. To check the reliability of generated masks, we use them for training segmentation models. Then we can compare the performance of models trained on generated mask with the performance of models trained on the GTs.
This experiment has been done for nuclear segmentation task, where we trained three well-known segmentation networks (U-Net \cite{ronneberger2015u}, SegNet \cite{badrinarayanan2017segnet}, and FCN8 \cite{long2015fully}) with GT and NuClick generated masks separately and evaluated the trained models on the validation set. Results of these experiments are reported in \cref{tRelaiable}. Note that when we are evaluating the segmentation on MoNuSeg dataset, the NuClick model that generated the masks is trained on the CPM dataset. Therefore, in that case NuClick framework did not see any of MoNuSeg images during its training.

As shown in \cref{tRelaiable} there is a negligible difference between the metrics achieved by models trained on GT masks and the ones that trained on NuClick generated masks. Even for one instance, when testing on MoNuSeg dataset, Dice and SQ values resulted from FCN8 model trained on annotations of NuClick\textsubscript{CPM} are 0.01 and 0.006 (insignificantly) higher than the model trained on GT annotations, respectively. This might be due to more uniformity of the NuClick generated annotations, which eliminate the negative effect of inter annotator variations present in GT annotations. Therefore, the dense annotations generated by NuClick are reliable enough for using in practice. If we consider the cost of manual annotation, it is more efficient to use annotations obtained from NuClick to train models.

\begin{table}
\centering
\setlength{\tabcolsep}{1pt}
\caption{Results of segmentation reliability experiments}
\begin{tabular}{lcc|cc|cc|cc}
\hline\hline
\multirow{3}{*}{} & \multicolumn{4}{c|}{Result on MoNuSeg test set}                                 & \multicolumn{4}{c}{Result on CPM test set}                                      \\ 
\cline{2-9}
                  & \multicolumn{2}{c|}{GT} & \multicolumn{2}{c|}{NuClick\textsubscript{CPM}} & \multicolumn{2}{c|}{GT} & \multicolumn{2}{c}{NuClick\textsubscript{MoNuSeg}}  \\ 
\cline{2-9}
                  & Dice & SQ               & Dice & SQ                        & Dice & SQ               & Dice & SQ                          \\ 
\hline
Unet              & 0.825 & 0.510           & 0.824 & 0.503                      & 0.862 & 0.596             & 0.854 & 0.584                        \\
SegNet            & 0.849 & 0.531           & 0.842 & 0.527                      & 0.889 & 0.644             & 0.881 & 0.632                        \\
FCN8              & 0.808 & 0.453           & 0.818 & 0.459                      & 0.848 & 0.609             & 0.836 & 0.603   \\
\hline\hline
\end{tabular}
\label{tRelaiable}
\end{table}

\begin{figure*}[h!]
\centering
\includegraphics[width =\textwidth]{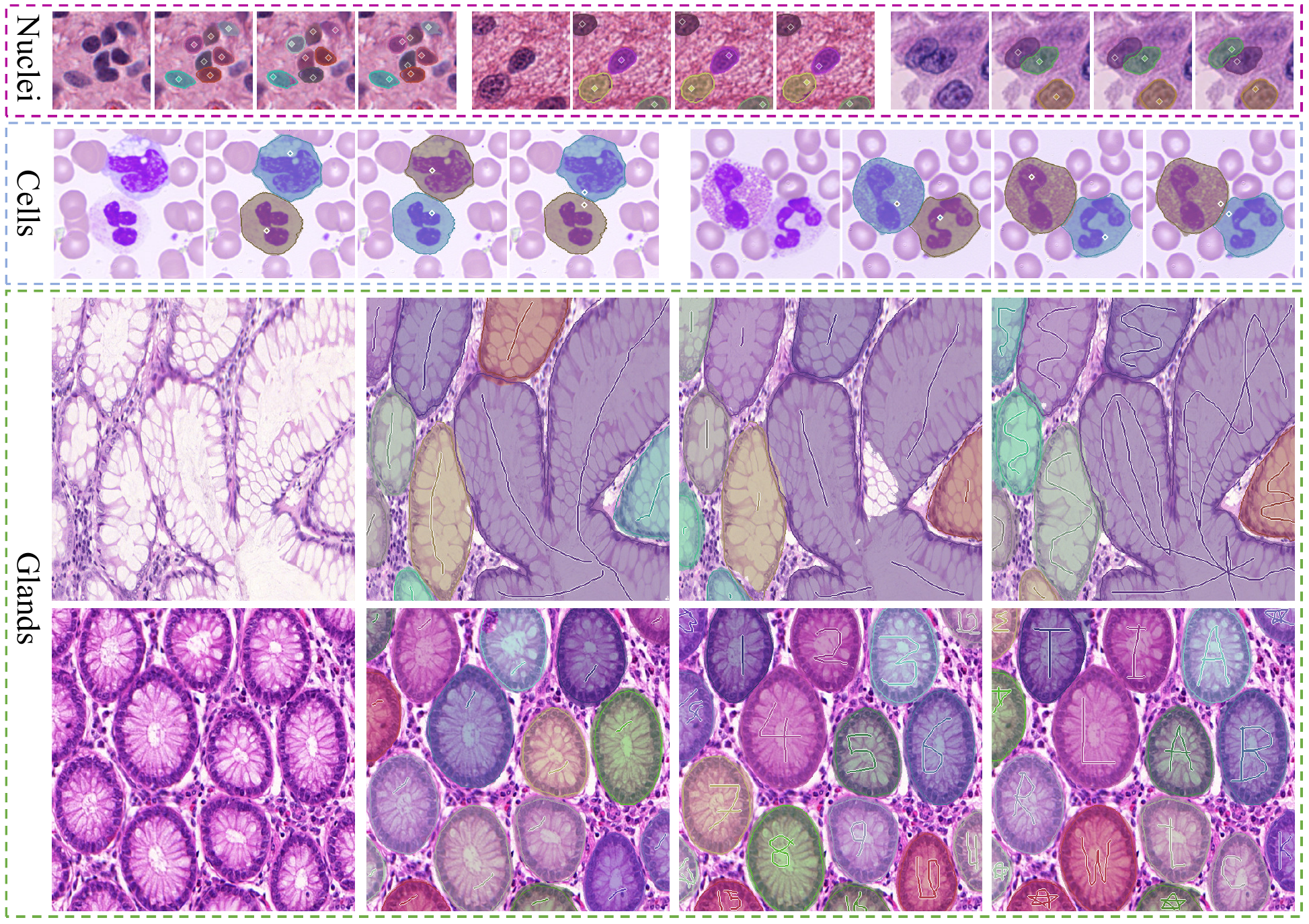}
\caption{Example results of NuClick, highlighting the variations in the user input. First and second rows show the prediction of Nuclick at different positions of clicks inside objects. The third and fourth rows demonstrates the predictions of nuclick in presense of variouse shape of squiggle. Solid stroke line around each object outlines the ground truth boundary for that object,  overlaid transparent mask is the predicted segmentation region by NuClick, and points or squiggles indicate the provided guiding signal for interactive segmentation. (Best viewed in color, zoom in to clearly see boundaries)}
\label{fig:sens}
\end{figure*}

\begin{figure*} [ht!]
\includegraphics[width =\textwidth]{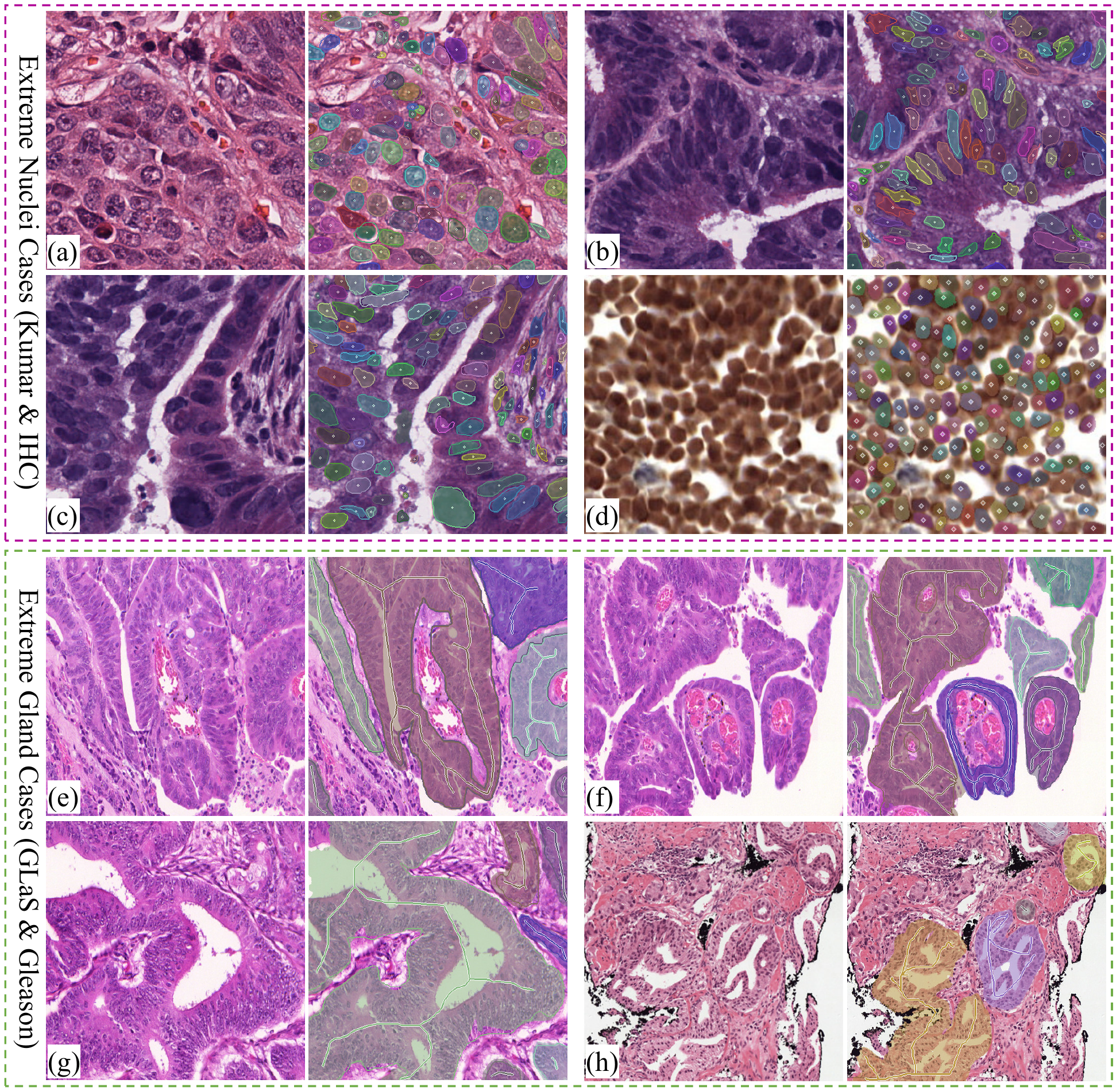}
Extreme cases for nuclei and glands: clumped nuclei in H\&E and IHC images (a-d) and irregular glands/tumor regions in cancerous colon and prostate images (e-h) are shown. In all images, solid stroke line around each object outlines the ground truth boundary for that object (except for d and e where the ground truth masks are unavailable),  overlaid transparent mask is the predicted segmentation region by NuClick, and points or squiggles indicate the provided guiding signal for interactive segmentation. (Best viewed in color, zoom in to clearly see boundaries)
\label{fig:extreme}
\end{figure*}

\begin{table}
\centering
\caption{Effect of disturbing click positions by  amount of $\sigma$ on NuClick outputs for nuclei and cells segmentation}
\label{tValsigma}
\begin{tabular}{lccc|ccc}
\hline\hline
                          & \multicolumn{3}{c|}{Nuclei} & \multicolumn{3}{c}{Cells (WBCs)}  \\ 
\hline
$\sigma$ & AJI   & Dice  & PQ.         & AJI   & Dice  & PQ.               \\ 
\hline
1                         & 0.834 & 0.912 & 0.838       & 0.954 & 0.983 & 0.958             \\
3                         & 0.834 & 0.911 & 0.837       & 0.954 & 0.983 & 0.958             \\
5                         & 0.832 & 0.911 & 0.835       & 0.953 & 0.983 & 0.957             \\
10                        & 0.821 & 0.903 & 0.822       & 0.953 & 0.982 & 0.957             \\
20                        & -     & -     & -           & 0.950 & 0.979 & 0.955             \\
50                        & -     & -     & -           & 0.935 & 0.961 & 0.943             \\
\hline\hline
\end{tabular}
\label{tSensitivity}
\end{table}
\subsection{Sensitivity to Guiding Signals}
Performance of an interactive segmentation algorithm highly depends on quality of the user input markers.
In other words, an ideal interactive segmentation tool must be robust against errors in the input annotations as much as possible.  
For instance, in nucleus or cell segmentation, an ideal segmentation tools should perform well to delineate boundaries of nuclei as long as user clicks fall inside the nuclei region i.e., the clicked point does not need to be located exactly at the center of the desired nuclei.
\begin{table}
\centering
\caption{Effect of disturbing click positions by  amount of $\sigma$ on NuClick outputs for nuclei and cells segmentation}
\label{tValsigma}
\begin{tabular}{lccc|ccc}
\hline\hline
                          & \multicolumn{3}{c|}{Nuclei} & \multicolumn{3}{c}{Cells (WBCs)}  \\ 
\hline
$\sigma$ & AJI   & Dice  & PQ.         & AJI   & Dice  & PQ.               \\ 
\hline
1                         & 0.834 & 0.912 & 0.838       & 0.954 & 0.983 & 0.958             \\
3                         & 0.834 & 0.911 & 0.837       & 0.954 & 0.983 & 0.958             \\
5                         & 0.832 & 0.911 & 0.835       & 0.953 & 0.983 & 0.957             \\
10                        & 0.821 & 0.903 & 0.822       & 0.953 & 0.982 & 0.957             \\
20                        & -     & -     & -           & 0.950 & 0.979 & 0.955             \\
50                        & -     & -     & -           & 0.935 & 0.961 & 0.943             \\
\hline\hline
\end{tabular}
\label{tSensitivity}
\end{table}
To assess the sensitivity of NuClick to the variations in the guiding signal, we design an experiment for nuclei and cell segmentation applications in which location of the guiding point in the inclusion map is perturbed by adding value of ${\sigma}$ to the location of centroids. We repeat this experiment for different values of $\sigma$ for both nuclei and cell segmentation applications and report the results in \cref{tSensitivity}. For nuclear segmentation, jittering the location up to 10 pixels is investigated. It has been shown that disturbing the click position from the centroid up to 5 pixels does not considerably degrade the segmentation results. However, when the jittering amount is equal to $\sigma=10$, all evaluation metrics drop by 1\% or more. This reduction in metrics does not necessarily imply that NuClick is sensitive to click positions, because this fall in performance may be due to the fact that radius of some nuclei is less than 10 pixels and jittering the click position by 10 pixels cause it to fall outside the nuclei region therefore confusing the NuClick in correctly segmenting the desired small nucleus. However, even reduced metrics are still reliable in comparison with the resulted metrics from other methods as reported in \cref{tValNuc}.

The same trend can be seen for cell segmentaiton task in \cref{tSensitivity}. However, for cells in our dataset we were able to increase the jittering range (up to 50 pixels) because in the WBC dataset, white blood cells have a diameter of at least 80 pixels. As one can see, the segmentation results are very robust against the applied distortion to the click position. Changing the click location by 50 pixels makes considerable drop in performance which can be due to the same reason as we discussed the nuclei case i.e., amount of jittering is bigger than the average radius of some small cells.

Unfortunately, we can not quantitatively analyze the sensitivity of the NuClick to the squiggle changes, because its related changes are not easily measurable/paramtereizable. However, for two examples of histology images we tried to show the effect of changing the guiding squiggles on the resulting segmentation in \cref{fig:sens}. In this figure, the effect of changing the click position for two examples of nuclei segmentation and two examples cell segmentation are also visualized. It is obvious from exemplars in \cref{fig:sens} that NuClick successfully works with different shapes of squiggles as the guiding signal. Squiggles can be short in the middle or adjacent regions of the desired gland, or they can be long enough to cover the main diameter of the gland. They can be continuous curves covering all section and indentation of the gland geometry, or separated discrete lines that indicate different sections of a big gland. They can even have arbitrary numerical or letters shape like the example in the last row of \cref{fig:sens}. In all cases, it is obvious that NuClick is quite robust against variations in  the guiding signals which is due to the techniques that we have incorporated during training of the NuClick (randomizing the inclusion map).

It is worth mentioning that we have conducted experiments with training NuClick for gland segmentation using extreme points and polygons as guiding signals. Even with a considerable number of points on gland boundary or  polygons with large number of vertices (filled or hollow), the network failed to converge during the training phase. However, we observed that even simple or small squiggles are able to provide enough guiding information for the model to converge fast.

We have also conducted another experiment to assess the sensitivity of NuClick on the exclusion maps. In other words, we want to see if eliminating the exclusion map has any effect on NuClick segmentation performance. To this end,  we  evaluate the performance of NuClick for nuclei segmentation on MoNuSeg dataset in the absence of exclusion map. Therefore in this situation the input to the network would have 4 channels (RGB plus inclusion map). The network is trained from scratch on the MoNuSeg training set with the new considerations and then evaluated on the MoNuSeg validation set. Results of this experiment are reported in \cref{tExclud}. Based on \cref{tExclud}, performance of the NuClick significantly drops when exclusion map is missing. That is because there are a lot of overlapping nuclei in this dataset and without having the exclusion map, the network has no clue of the neighboring nuclei when dealing with a nucleus that belongs to a nuclei clump.

\subsection{Extreme Cases}
To investigate the effectiveness of NuClick when dealing with extreme cases, output of NuClick for images with challenging objects (high grade cancer in different tissue types)  are shown in \cref{fig:extreme}. For example in \cref{fig:extreme}a-c touching nuclei with unclear edges from patches of cancerous samples have been successfully segmented by NuClick. Additionally, \cref{fig:extreme}d shows promising segmentation of densely clustered blood cells in  a blurred IHC image from another domain (extracted from LYON19 dataset (\cite{lyon19})).

In \cref{fig:extreme}e-f, images of glands with irregular shapes  and their overlaid predictions are shown. As long as the squiggle covers the extend of gland, we can achieve a good segmentation.
A noteworthy property of NuClick framework is its capability to segment objects with holes in them. In \cref{fig:extreme}e-f, although margins of glands are very unclear and some  glands have holes in their shape,   NuClick can successfully recognizing boundaries of each gland. Further, 
if the squiggle encompass the hole, it will be excluded from final segmentation whereas if the squiggle covers part of holes in the middle of glands, they will be included in the segmentation. For instance, in \cref{fig:extreme}g, a complex and relatively large gland is well delineated by the NuClick. Note that this gland contains a hole region which belongs to the gland and it is correctly segmented as part of the gland because the guiding signal covers that part.
This is a powerful and very useful property that methods based on extreme points or bounding box like \cite{maninis2018deep} and \cite{wang2018interactive} do not offer.

We also show a cancerous prostate image (extracted from PANDA dataset (\cite{bulten2020automated})) in \cref{fig:extreme}h where the tumor regions are outlined by NuClick. Overall, these predictions shows the capability of  NuClick in providing reasonable annotation in scenarios that are even challenging for humans to annotate. Note that for images in \cref{fig:extreme}d,h the ground truth segmentation masks are not available, therefore they are not shown.

\begin{table}
\centering
\setlength{\tabcolsep}{2.5pt}
\caption{Performance of NuClick on the MonuSeg dataset with and without exclusion map}
\begin{tabular}{llllll}
                        & AJI   & Dice  & SQ    & DQ    & PQ     \\ 
\hline\hline
NuClick with ex. map    & 0.834 & 0.912 & 0.839 & 0.999 & 0.838  \\
NuClick without ex. map & 0.815 & 0.894 & 0.801 & 0.972 & 0.778 
\end{tabular}
\label{tExclud}
\end{table}

\subsection{User Correction}
In some cases, the output of models might not be correct, therefore there should be a possibility that user can modify wrong predictions.  This is a matter of implementation of the interface in most cases, Hence,  when the output is not as good as expected, the user can modify the supervisory signal by extending squiggles,  changing the shape of squiggles or move the position of clicks. After the modification has been applied, the new modified supervisory signal is fed to the network to obtain new segmentation.

\section{Conclusions}
In this paper, we have presented NuClick, a CNN-based framework for interactive segmentation of objects in histology images. We proposed a simple and robust way to provide input from the user which  minimizes human effort for obtaining dense annotations of nuclei, cell and glands in histology. We showed that our method is generizable enough to be used across different datasets and it can be used even for annotating objects from completely different data distributions.  Applicability of NuClick has been shown across 6 datasets, where NuClick obtained state-of-the art performance in all scenarios. NuClick can also be used for segmenting other objects like nerves and vessels which are less complex and less heterogeneous compared to glands. We believe that NuClick can be used as a useful plug-in for whole slide annotation programs like ASAP (\cite{litjens2017asap}) or Qupath (\cite{bankhead2017qupath}) to ease the labeling process of the large-scale datasets.

\bibliographystyle{unsrt}
\bibliography{refs.bib}

\begin{thebibliography}{10}

\bibitem{russakovsky2015imagenet}
Olga Russakovsky, Jia Deng, Hao Su, Jonathan Krause, Sanjeev Satheesh, Sean Ma,
  Zhiheng Huang, Andrej Karpathy, Aditya Khosla, Michael Bernstein, et~al.
\newblock Imagenet large scale visual recognition challenge.
\newblock {\em International journal of computer vision}, 115(3):211--252,
  2015.

\bibitem{taghanaki2019deep}
Saed~Asgari Taghanaki, Kumar Abhishek, Joseph~Paul Cohen, Julien Cohen-Adad,
  and Ghassan Hamarneh.
\newblock Deep semantic segmentation of natural and medical images: A review.
\newblock {\em accepted to appear in Springer Artificial Intelligence Review},
  2020.

\bibitem{sirinukunwattana2017gland}
Korsuk Sirinukunwattana, Josien~PW Pluim, Hao Chen, Xiaojuan Qi, Pheng-Ann
  Heng, Yun~Bo Guo, Li~Yang Wang, Bogdan~J Matuszewski, Elia Bruni, Urko
  Sanchez, et~al.
\newblock Gland segmentation in colon histology images: The glas challenge
  contest.
\newblock {\em Medical image analysis}, 35:489--502, 2017.

\bibitem{kumar2019multi}
Neeraj Kumar, Ruchika Verma, Deepak Anand, Yanning Zhou, Omer~Fahri Onder,
  Efstratios Tsougenis, Hao Chen, Pheng~Ann Heng, Jiahui Li, Zhiqiang Hu,
  et~al.
\newblock A multi-organ nucleus segmentation challenge.
\newblock {\em IEEE transactions on medical imaging}, 2019.

\bibitem{graham2019hover}
Simon Graham, Quoc~Dang Vu, Shan E~Ahmed Raza, Ayesha Azam, Yee~Wah Tsang,
  Jin~Tae Kwak, and Nasir Rajpoot.
\newblock Hover-net: Simultaneous segmentation and classification of nuclei in
  multi-tissue histology images.
\newblock {\em Medical Image Analysis}, 58:101563, 2019.

\bibitem{koohbanani2019nuclear}
Navid~Alemi Koohbanani, Mostafa Jahanifar, Ali Gooya, and Nasir Rajpoot.
\newblock Nuclear instance segmentation using a proposal-free spatially aware
  deep learning framework.
\newblock In {\em International Conference on Medical Image Computing and
  Computer-Assisted Intervention}, pages 622--630. Springer, 2019.

\bibitem{pinckaers2019neural}
Hans Pinckaers and Geert Litjens.
\newblock Neural ordinary differential equations for semantic segmentation of
  individual colon glands.
\newblock {\em arXiv preprint arXiv:1910.10470}, 2019.

\bibitem{graham2019mild}
Simon Graham, Hao Chen, Jevgenij Gamper, Qi~Dou, Pheng-Ann Heng, David Snead,
  Yee~Wah Tsang, and Nasir Rajpoot.
\newblock Mild-net: Minimal information loss dilated network for gland instance
  segmentation in colon histology images.
\newblock {\em Medical image analysis}, 52:199--211, 2019.

\bibitem{chen2016dcan}
Hao Chen, Xiaojuan Qi, Lequan Yu, and Pheng-Ann Heng.
\newblock Dcan: deep contour-aware networks for accurate gland segmentation.
\newblock In {\em Proceedings of the IEEE conference on Computer Vision and
  Pattern Recognition}, pages 2487--2496, 2016.

\bibitem{gamper2020pannuke}
Jevgenij Gamper, Navid~Alemi Koohbanani, Simon Graham, Mostafa Jahanifar,
  Syed~Ali Khurram, Ayesha Azam, Katherine Hewitt, and Nasir Rajpoot.
\newblock Pannuke dataset extension, insights and baselines.
\newblock {\em arXiv preprint arXiv:2003.10778}, 2020.

\bibitem{zhou2019cia}
Yanning Zhou, Omer~Fahri Onder, Qi~Dou, Efstratios Tsougenis, Hao Chen, and
  Pheng-Ann Heng.
\newblock Cia-net: Robust nuclei instance segmentation with contour-aware
  information aggregation.
\newblock In {\em International Conference on Information Processing in Medical
  Imaging}, pages 682--693. Springer, 2019.

\bibitem{yoo2019pseudoedgenet}
Inwan Yoo, Donggeun Yoo, and Kyunghyun Paeng.
\newblock Pseudoedgenet: Nuclei segmentation only with point annotations.
\newblock In {\em International Conference on Medical Image Computing and
  Computer-Assisted Intervention}, pages 731--739. Springer, 2019.

\bibitem{qu2019weakly}
Hui Qu, Pengxiang Wu, Qiaoying Huang, Jingru Yi, Gregory~M Riedlinger,
  Subhajyoti De, and Dimitris~N Metaxas.
\newblock Weakly supervised deep nuclei segmentation using points annotation in
  histopathology images.
\newblock In {\em International Conference on Medical Imaging with Deep
  Learning}, pages 390--400, 2019.

\bibitem{pathak2014fully}
Deepak Pathak, Evan Shelhamer, Jonathan Long, and Trevor Darrell.
\newblock Fully convolutional multi-class multiple instance learning.
\newblock {\em arXiv preprint arXiv:1412.7144}, 2014.

\bibitem{kolesnikov2016seed}
Alexander Kolesnikov and Christoph~H Lampert.
\newblock Seed, expand and constrain: Three principles for weakly-supervised
  image segmentation.
\newblock In {\em European Conference on Computer Vision}, pages 695--711.
  Springer, 2016.

\bibitem{pathak2015constrained}
Deepak Pathak, Philipp Krahenbuhl, and Trevor Darrell.
\newblock Constrained convolutional neural networks for weakly supervised
  segmentation.
\newblock In {\em Proceedings of the IEEE international conference on computer
  vision}, pages 1796--1804, 2015.

\bibitem{wei2018revisiting}
Yunchao Wei, Huaxin Xiao, Honghui Shi, Zequn Jie, Jiashi Feng, and Thomas~S
  Huang.
\newblock Revisiting dilated convolution: A simple approach for weakly-and
  semi-supervised semantic segmentation.
\newblock In {\em Proceedings of the IEEE Conference on Computer Vision and
  Pattern Recognition}, pages 7268--7277, 2018.

\bibitem{khoreva2017simple}
Anna Khoreva, Rodrigo Benenson, Jan Hosang, Matthias Hein, and Bernt Schiele.
\newblock Simple does it: Weakly supervised instance and semantic segmentation.
\newblock In {\em Proceedings of the IEEE conference on computer vision and
  pattern recognition}, pages 876--885, 2017.

\bibitem{jin2017webly}
Bin Jin, Maria~V Ortiz~Segovia, and Sabine Susstrunk.
\newblock Webly supervised semantic segmentation.
\newblock In {\em Proceedings of the IEEE Conference on Computer Vision and
  Pattern Recognition}, pages 3626--3635, 2017.

\bibitem{ahmed2014semantic}
Ejaz Ahmed, Scott Cohen, and Brian Price.
\newblock Semantic object selection.
\newblock In {\em Proceedings of the IEEE Conference on Computer Vision and
  Pattern Recognition}, pages 3150--3157, 2014.

\bibitem{bearman2016s}
Amy Bearman, Olga Russakovsky, Vittorio Ferrari, and Li~Fei-Fei.
\newblock What’s the point: Semantic segmentation with point supervision.
\newblock In {\em European conference on computer vision}, pages 549--565.
  Springer, 2016.

\bibitem{bell2015material}
Sean Bell, Paul Upchurch, Noah Snavely, and Kavita Bala.
\newblock Material recognition in the wild with the materials in context
  database.
\newblock In {\em Proceedings of the IEEE conference on computer vision and
  pattern recognition}, pages 3479--3487, 2015.

\bibitem{chen2018tap}
Ding-Jie Chen, Jui-Ting Chien, Hwann-Tzong Chen, and Long-Wen Chang.
\newblock Tap and shoot segmentation.
\newblock In {\em Thirty-Second AAAI Conference on Artificial Intelligence},
  2018.

\bibitem{wang2014touchcut}
Tinghuai Wang, Bo~Han, and John Collomosse.
\newblock Touchcut: Fast image and video segmentation using single-touch
  interaction.
\newblock {\em Computer Vision and Image Understanding}, 120:14--30, 2014.

\bibitem{lin2016scribblesup}
Di~Lin, Jifeng Dai, Jiaya Jia, Kaiming He, and Jian Sun.
\newblock Scribblesup: Scribble-supervised convolutional networks for semantic
  segmentation.
\newblock In {\em Proceedings of the IEEE Conference on Computer Vision and
  Pattern Recognition}, pages 3159--3167, 2016.

\bibitem{xu2015learning}
Jia Xu, Alexander~G Schwing, and Raquel Urtasun.
\newblock Learning to segment under various forms of weak supervision.
\newblock In {\em Proceedings of the IEEE conference on computer vision and
  pattern recognition}, pages 3781--3790, 2015.

\bibitem{bai2009geodesic}
Xue Bai and Guillermo Sapiro.
\newblock Geodesic matting: A framework for fast interactive image and video
  segmentation and matting.
\newblock {\em International journal of computer vision}, 82(2):113--132, 2009.

\bibitem{batra2011interactively}
Dhruv Batra, Adarsh Kowdle, Devi Parikh, Jiebo Luo, and Tsuhan Chen.
\newblock Interactively co-segmentating topically related images with
  intelligent scribble guidance.
\newblock {\em International journal of computer vision}, 93(3):273--292, 2011.

\bibitem{boykov2001interactive}
Yuri~Y Boykov and M-P Jolly.
\newblock Interactive graph cuts for optimal boundary \& region segmentation of
  objects in nd images.
\newblock In {\em Proceedings eighth IEEE international conference on computer
  vision. ICCV 2001}, volume~1, pages 105--112. IEEE, 2001.

\bibitem{rother2004grabcut}
Carsten Rother, Vladimir Kolmogorov, and Andrew Blake.
\newblock Grabcut: Interactive foreground extraction using iterated graph cuts.
\newblock In {\em ACM transactions on graphics (TOG)}, volume~23, pages
  309--314. ACM, 2004.

\bibitem{cheng2015densecut}
Ming-Ming Cheng, Victor~Adrian Prisacariu, Shuai Zheng, Philip~HS Torr, and
  Carsten Rother.
\newblock Densecut: Densely connected crfs for realtime grabcut.
\newblock In {\em Computer Graphics Forum}, volume~34, pages 193--201. Wiley
  Online Library, 2015.

\bibitem{gulshan2010geodesic}
Varun Gulshan, Carsten Rother, Antonio Criminisi, Andrew Blake, and Andrew
  Zisserman.
\newblock Geodesic star convexity for interactive image segmentation.
\newblock In {\em 2010 IEEE Computer Society Conference on Computer Vision and
  Pattern Recognition}, pages 3129--3136. IEEE, 2010.

\bibitem{shankar2015video}
Naveen Shankar~Nagaraja, Frank~R Schmidt, and Thomas Brox.
\newblock Video segmentation with just a few strokes.
\newblock In {\em Proceedings of the IEEE International Conference on Computer
  Vision}, pages 3235--3243, 2015.

\bibitem{mortensen1998interactive}
Eric~N Mortensen and William~A Barrett.
\newblock Interactive segmentation with intelligent scissors.
\newblock {\em Graphical models and image processing}, 60(5):349--384, 1998.

\bibitem{cagnoni1999genetic}
Stefano Cagnoni, Andrew~B Dobrzeniecki, Riccardo Poli, and Jacquelyn~C Yanch.
\newblock Genetic algorithm-based interactive segmentation of 3d medical
  images.
\newblock {\em Image and Vision Computing}, 17(12):881--895, 1999.

\bibitem{de2004interactive}
Marleen de~Bruijne, Bram van Ginneken, Max~A Viergever, and Wiro~J Niessen.
\newblock Interactive segmentation of abdominal aortic aneurysms in cta images.
\newblock {\em Medical Image Analysis}, 8(2):127--138, 2004.

\bibitem{wang2018interactive}
Guotai Wang, Wenqi Li, Maria~A Zuluaga, Rosalind Pratt, Premal~A Patel, Michael
  Aertsen, Tom Doel, Anna~L David, Jan Deprest, S{\'e}bastien Ourselin, et~al.
\newblock Interactive medical image segmentation using deep learning with
  image-specific fine tuning.
\newblock {\em IEEE transactions on medical imaging}, 37(7):1562--1573, 2018.

\bibitem{li2018interactive}
Zhuwen Li, Qifeng Chen, and Vladlen Koltun.
\newblock Interactive image segmentation with latent diversity.
\newblock In {\em Proceedings of the IEEE Conference on Computer Vision and
  Pattern Recognition}, pages 577--585, 2018.

\bibitem{papadopoulos2017extreme}
Dim~P Papadopoulos, Jasper~RR Uijlings, Frank Keller, and Vittorio Ferrari.
\newblock Extreme clicking for efficient object annotation.
\newblock In {\em Proceedings of the IEEE International Conference on Computer
  Vision}, pages 4930--4939, 2017.

\bibitem{kwatra2003graphcut}
Vivek Kwatra, Arno Sch{\"o}dl, Irfan Essa, Greg Turk, and Aaron Bobick.
\newblock Graphcut textures: image and video synthesis using graph cuts.
\newblock In {\em ACM Transactions on Graphics (ToG)}, volume~22, pages
  277--286. ACM, 2003.

\bibitem{xu2017deep}
Ning Xu, Brian Price, Scott Cohen, Jimei Yang, and Thomas Huang.
\newblock Deep grabcut for object selection.
\newblock {\em arXiv preprint arXiv:1707.00243}, 2017.

\bibitem{xu2016deep}
Ning Xu, Brian Price, Scott Cohen, Jimei Yang, and Thomas~S Huang.
\newblock Deep interactive object selection.
\newblock In {\em Proceedings of the IEEE Conference on Computer Vision and
  Pattern Recognition}, pages 373--381, 2016.

\bibitem{agustsson2019interactive}
Eirikur Agustsson, Jasper~RR Uijlings, and Vittorio Ferrari.
\newblock Interactive full image segmentation by considering all regions
  jointly.
\newblock In {\em Proceedings of the IEEE Conference on Computer Vision and
  Pattern Recognition}, pages 11622--11631, 2019.

\bibitem{maninis2018deep}
Kevis-Kokitsi Maninis, Sergi Caelles, Jordi Pont-Tuset, and Luc Van~Gool.
\newblock Deep extreme cut: From extreme points to object segmentation.
\newblock In {\em Proceedings of the IEEE Conference on Computer Vision and
  Pattern Recognition}, pages 616--625, 2018.

\bibitem{ling2019fast}
Huan Ling, Jun Gao, Amlan Kar, Wenzheng Chen, and Sanja Fidler.
\newblock Fast interactive object annotation with curve-gcn.
\newblock In {\em Proceedings of the IEEE Conference on Computer Vision and
  Pattern Recognition}, pages 5257--5266, 2019.

\bibitem{castrejon2017annotating}
Lluis Castrejon, Kaustav Kundu, Raquel Urtasun, and Sanja Fidler.
\newblock Annotating object instances with a polygon-rnn.
\newblock In {\em Proceedings of the IEEE Conference on Computer Vision and
  Pattern Recognition}, pages 5230--5238, 2017.

\bibitem{acuna2018efficient}
David Acuna, Huan Ling, Amlan Kar, and Sanja Fidler.
\newblock Efficient interactive annotation of segmentation datasets with
  polygon-rnn++.
\newblock In {\em Proceedings of the IEEE Conference on Computer Vision and
  Pattern Recognition}, pages 859--868, 2018.

\bibitem{wang2019object}
Zian Wang, David Acuna, Huan Ling, Amlan Kar, and Sanja Fidler.
\newblock Object instance annotation with deep extreme level set evolution.
\newblock In {\em Proceedings of the IEEE Conference on Computer Vision and
  Pattern Recognition}, pages 7500--7508, 2019.

\bibitem{caselles1997geodesic}
Vicent Caselles, Ron Kimmel, and Guillermo Sapiro.
\newblock Geodesic active contours.
\newblock {\em International journal of computer vision}, 22(1):61--79, 1997.

\bibitem{acuna2019devil}
David Acuna, Amlan Kar, and Sanja Fidler.
\newblock Devil is in the edges: Learning semantic boundaries from noisy
  annotations.
\newblock In {\em Proceedings of the IEEE Conference on Computer Vision and
  Pattern Recognition}, pages 11075--11083, 2019.

\bibitem{sakinis2019interactive}
Tomas Sakinis, Fausto Milletari, Holger Roth, Panagiotis Korfiatis, Petro
  Kostandy, Kenneth Philbrick, Zeynettin Akkus, Ziyue Xu, Daguang Xu, and
  Bradley~J Erickson.
\newblock Interactive segmentation of medical images through fully
  convolutional neural networks.
\newblock {\em arXiv preprint arXiv:1903.08205}, 2019.

\bibitem{andriluka2018fluid}
Mykhaylo Andriluka, Jasper~RR Uijlings, and Vittorio Ferrari.
\newblock Fluid annotation: a human-machine collaboration interface for full
  image annotation.
\newblock In {\em Proceedings of the 26th ACM international conference on
  Multimedia}, pages 1957--1966, 2018.

\bibitem{he2017mask}
Kaiming He, Georgia Gkioxari, Piotr Doll{\'a}r, and Ross Girshick.
\newblock Mask r-cnn.
\newblock In {\em Proceedings of the IEEE international conference on computer
  vision}, pages 2961--2969, 2017.

\bibitem{nieuwenhuis2012spatially}
Claudia Nieuwenhuis and Daniel Cremers.
\newblock Spatially varying color distributions for interactive multilabel
  segmentation.
\newblock {\em IEEE transactions on pattern analysis and machine intelligence},
  35(5):1234--1247, 2012.

\bibitem{nieuwenhuis2014co}
Claudia Nieuwenhuis, Simon Hawe, Martin Kleinsteuber, and Daniel Cremers.
\newblock Co-sparse textural similarity for interactive segmentation.
\newblock In {\em European conference on computer vision}, pages 285--301.
  Springer, 2014.

\bibitem{santner2010interactive}
Jakob Santner, Thomas Pock, and Horst Bischof.
\newblock Interactive multi-label segmentation.
\newblock In {\em Asian Conference on Computer Vision}, pages 397--410.
  Springer, 2010.

\bibitem{vezhnevets2005growcut}
Vladimir Vezhnevets and Vadim Konouchine.
\newblock Growcut: Interactive multi-label nd image segmentation by cellular
  automata.
\newblock In {\em proc. of Graphicon}, volume~1, pages 150--156. Citeseer,
  2005.

\bibitem{jahanifar2019nuclick}
Mostafa Jahanifar, Navid~Alemi Koohbanani, and Nasir Rajpoot.
\newblock Nuclick: From clicks in the nuclei to nuclear boundaries.
\newblock {\em arXiv preprint arXiv:1909.03253}, 2019.

\bibitem{wu2014milcut}
Jiajun Wu, Yibiao Zhao, Jun-Yan Zhu, Siwei Luo, and Zhuowen Tu.
\newblock Milcut: A sweeping line multiple instance learning paradigm for
  interactive image segmentation.
\newblock In {\em Proceedings of the IEEE Conference on Computer Vision and
  Pattern Recognition}, pages 256--263, 2014.

\bibitem{rother2012interactive}
Carsten Rother, Vladimir Kolmogorov, and Andrew Blake.
\newblock Interactive foreground extraction using iterated graph cuts.
\newblock {\em ACM Transactions on Graphics}, (23):3, 2012.

\bibitem{hesamian2019deep}
Mohammad~Hesam Hesamian, Wenjing Jia, Xiangjian He, and Paul Kennedy.
\newblock Deep learning techniques for medical image segmentation: Achievements
  and challenges.
\newblock {\em Journal of digital imaging}, 32(4):582--596, 2019.

\bibitem{garcia2017review}
Alberto Garcia-Garcia, Sergio Orts-Escolano, Sergiu Oprea, Victor
  Villena-Martinez, and Jose Garcia-Rodriguez.
\newblock A review on deep learning techniques applied to semantic
  segmentation.
\newblock {\em arXiv preprint arXiv:1704.06857}, 2017.

\bibitem{jahanifar2018segmentation}
Mostafa Jahanifar, Neda~Zamani Tajeddin, Navid~Alemi Koohbanani, Ali Gooya, and
  Nasir Rajpoot.
\newblock Segmentation of skin lesions and their attributes using multi-scale
  convolutional neural networks and domain specific augmentations.
\newblock {\em arXiv preprint arXiv:1809.10243}, 2018.

\bibitem{chen2017rethinking}
Liang-Chieh Chen, George Papandreou, Florian Schroff, and Hartwig Adam.
\newblock Rethinking atrous convolution for semantic image segmentation.
\newblock {\em arXiv preprint arXiv:1706.05587}, 2017.

\bibitem{he2016deep}
Kaiming He, Xiangyu Zhang, Shaoqing Ren, and Jian Sun.
\newblock Deep residual learning for image recognition.
\newblock In {\em Proceedings of the IEEE conference on computer vision and
  pattern recognition}, pages 770--778, 2016.

\bibitem{serra1983image}
Jean Serra.
\newblock {\em Image analysis and mathematical morphology}.
\newblock Academic Press, Inc., 1983.

\bibitem{awan2017glandular}
Ruqayya Awan, Korsuk Sirinukunwattana, David Epstein, Samuel Jefferyes, Uvais
  Qidwai, Zia Aftab, Imaad Mujeeb, David Snead, and Nasir Rajpoot.
\newblock Glandular morphometrics for objective grading of colorectal
  adenocarcinoma histology images.
\newblock {\em Scientific reports}, 7(1):16852, 2017.

\bibitem{vu2019methods}
Quoc~Dang Vu, Simon Graham, Tahsin Kurc, Minh Nguyen~Nhat To, Muhammad Shaban,
  Talha Qaiser, Navid~Alemi Koohbanani, Syed~Ali Khurram, Jayashree
  Kalpathy-Cramer, Tianhao Zhao, et~al.
\newblock Methods for segmentation and classification of digital microscopy
  tissue images.
\newblock {\em Frontiers in bioengineering and biotechnology}, 7, 2019.

\bibitem{kumar2017dataset}
Neeraj Kumar, Ruchika Verma, Sanuj Sharma, Surabhi Bhargava, Abhishek Vahadane,
  and Amit Sethi.
\newblock A dataset and a technique for generalized nuclear segmentation for
  computational pathology.
\newblock {\em IEEE transactions on medical imaging}, 36(7):1550--1560, 2017.

\bibitem{kirillov2019panoptic}
Alexander Kirillov, Kaiming He, Ross Girshick, Carsten Rother, and Piotr
  Doll{\'a}r.
\newblock Panoptic segmentation.
\newblock In {\em Proceedings of the IEEE conference on computer vision and
  pattern recognition}, pages 9404--9413, 2019.

\bibitem{ronneberger2015u}
Olaf Ronneberger, Philipp Fischer, and Thomas Brox.
\newblock U-net: Convolutional networks for biomedical image segmentation.
\newblock In {\em International Conference on Medical image computing and
  computer-assisted intervention}, pages 234--241. Springer, 2015.

\bibitem{badrinarayanan2017segnet}
Vijay Badrinarayanan, Alex Kendall, and Roberto Cipolla.
\newblock Segnet: A deep convolutional encoder-decoder architecture for image
  segmentation.
\newblock {\em IEEE transactions on pattern analysis and machine intelligence},
  39(12):2481--2495, 2017.

\bibitem{long2015fully}
Jonathan Long, Evan Shelhamer, and Trevor Darrell.
\newblock Fully convolutional networks for semantic segmentation.
\newblock In {\em Proceedings of the IEEE conference on computer vision and
  pattern recognition}, pages 3431--3440, 2015.

\bibitem{adams1994seeded}
Rolf Adams and Leanne Bischof.
\newblock Seeded region growing.
\newblock {\em IEEE Transactions on pattern analysis and machine intelligence},
  16(6):641--647, 1994.

\bibitem{chan2001active}
Tony~F Chan and Luminita~A Vese.
\newblock Active contours without edges.
\newblock {\em IEEE Transactions on image processing}, 10(2):266--277, 2001.

\bibitem{parvati2008image}
K~Parvati, Prakasa Rao, and M~Mariya~Das.
\newblock Image segmentation using gray-scale morphology and marker-controlled
  watershed transformation.
\newblock {\em Discrete Dynamics in Nature and Society}, 2008, 2008.

\bibitem{lyon19}
Zaneta Swiderska-Chadaj, Hans Pinckaers, Mart van Rijthoven, Maschenka
  Balkenhol, Margarita Melnikova, Oscar Geessink, Quirine Manson, Mark Sherman,
  Antonio Polonia, Jeremy Parry, Mustapha Abubakar, Geert Litjens, Jeroen
  van~der Laak, and Francesco Ciompi.
\newblock Learning to detect lymphocytes in immunohistochemistry with deep
  learning.
\newblock {\em Medical Image Analysis}, 58:101547, 2019.

\bibitem{bulten2020automated}
Wouter Bulten, Hans Pinckaers, Hester van Boven, Robert Vink, Thomas de~Bel,
  Bram van Ginneken, Jeroen van~der Laak, Christina Hulsbergen-van~de Kaa, and
  Geert Litjens.
\newblock Automated deep-learning system for gleason grading of prostate cancer
  using biopsies: a diagnostic study.
\newblock {\em The Lancet Oncology}, 2020.

\bibitem{litjens2017asap}
G~Litjens.
\newblock Automated slide analysis platform (asap).
\newblock \url{http://rse.diagnijmegen.nl/software/asap/}, 2017.

\bibitem{bankhead2017qupath}
Peter Bankhead, Maurice~B Loughrey, Jos{\'e}~A Fern{\'a}ndez, Yvonne
  Dombrowski, Darragh~G McArt, Philip~D Dunne, Stephen McQuaid, Ronan~T Gray,
  Liam~J Murray, Helen~G Coleman, et~al.
\newblock Qupath: Open source software for digital pathology image analysis.
\newblock {\em Scientific reports}, 7(1):1--7, 2017.

\end{thebibliography}




\end{document}